\documentclass{article}
\pdfoutput=1

\PassOptionsToPackage{numbers}{natbib}

\usepackage[final]{neurips_2020}

\usepackage[utf8]{inputenc} 
\usepackage[T1]{fontenc}    
\usepackage{url}            
\usepackage{booktabs}       
\usepackage{amsfonts}       
\usepackage{nicefrac}       
\usepackage{color}
\usepackage{hyperref}       

\usepackage{amsthm}
\usepackage{amsmath}
\usepackage{float}
\usepackage{graphicx}
\usepackage{grffile}
\usepackage{wrapfig}
\usepackage{makecell}

\usepackage{caption}
\usepackage{subcaption}
\usepackage{bm}
\usepackage{multirow}
\usepackage{microtype}
\usepackage{mathtools}
\usepackage{tikz}
\usepackage{gensymb}
\usepackage{bbm}
\usepackage{amssymb}
\usepackage{enumitem}
\usepackage{placeins}
\usepackage{algorithm, algorithmic}
\usepackage{xspace}
\usepackage{rotating}
\usepackage{adjustbox}
\usepackage{diagbox}

\newcolumntype{R}[2]{%
    >{\adjustbox{angle=#1,lap=\width-(#2)}\bgroup}%
    l%
    <{\egroup}%
}

\renewcommand{\vec}[1]{\boldsymbol{\mathbf{#1}}}
\def\eg{{\em e.g.,}\xspace}
\def\ie{{\em i.e.,}\xspace}

\newcommand{\one}[1]{\mathbbm{1}_{[#1]}}
\def\expected{\mathbb{E}}
\def\t{T}
\def\bmu{\bm \mu}

\def\x{\vec{x}}
\def\z{\vec{z}}

\newcommand{\figref}[1]{Figure~\ref{#1}}

\newcommand{\tabref}[1]{Table~\ref{#1}}

\newcommand{\secref}[1]{Sec.~\ref{#1}}
\newcommand{\comment}[1]

\title{Exemplar VAE:~~~Linking Generative Models,\\Nearest Neighbor Retrieval, and Data Augmentation }

\author{
\begin{tabular}{ccc}
Sajad Norouzi$^{1,2}$ & David J.\ Fleet$^{1,2,3}$ & Mohamamd Norouzi$^{3}$  \\
\texttt{sajadn@cs.toronto.edu} & \texttt{fleet@cs.toronto.edu} & \texttt{mnorouzi@google.com}\\ 
\end{tabular}\\
${\ }^{1}${\normalfont University of Toronto},  
${\ }^{2}${\normalfont Vector Institute}, 
${\ }^{3}${\normalfont Google Research}\\

}

\begin{document}

\maketitle

\vspace{-.4cm}
\begin{abstract}
\vspace{-.3cm}
We introduce Exemplar VAEs, a family of generative models that bridge the
gap between {\em parametric} and {\em non-parametric, exemplar based} generative models.
Exemplar VAE is a variant of VAE with a non-parametric prior in the latent space based on a Parzen window estimator.
To sample from it, one first draws a random exemplar from a training set, then stochastically transforms that exemplar into a latent code and a new observation.
We propose retrieval augmented training (RAT) as a way to speed up Exemplar VAE training
by using approximate nearest neighbor search in the latent space to define a lower bound on log marginal likelihood.
To enhance generalization, model parameters are learned using exemplar leave-one-out and subsampling.
Experiments demonstrate the effectiveness of Exemplar VAEs on density estimation and representation learning. 
Importantly, generative data augmentation using Exemplar VAEs on permutation invariant MNIST and 
Fashion MNIST reduces classification error from 1.17\% to 0.69\% and from 8.56\% to 8.16\%. 
Code is available at \href{https://github.com/sajadn/Exemplar-VAE}{https://github.com/sajadn/Exemplar-VAE}.




\vspace{-.2cm}
\end{abstract}
\vspace{-.2cm}
\section{Introduction}
\vspace{-.1cm}



{\em Non-parametric, exemplar based}  methods use large, diverse sets of exemplars, and 
relatively simple learning algorithms such as Parzen window estimation~\cite{parzen1962estimation} 
and CRFs \cite{lafferty2001conditional}, to deliver impressive results on image generation 
(\eg~texture synthesis~\cite{efros1999texture}, image super resolution \cite{freeman2002example}, 
and inpaiting~\cite{criminisi2003object,hays2007scene}). 
These approaches generate new images by randomly selecting an exemplar 
from an existing dataset, and modifying it to form a new observation.
Sample quality of such models improves as dataset size increases, 
and additional training data can be  incorporated easily without further optimization.
However, exemplar based methods require a distance metric to define neighborhood structures,
and metric learning in high dimensional spaces is a challenge in itself~\cite{johnson2016perceptual,xing2003distance}.

Conversely, conventional {\em parametric} generative models based on deep neural nets enable learning
complex distributions (\eg~\cite{oord2016wavenet, reed2016generative}).
One can use standard generative frameworks~\cite{dinh2014nice, dinh2016density,goodfellow2014generative,
kingma2013auto,rezende2014stochastic} to optimize a decoder network to convert noise samples drawn from 
a factored Gaussian distribution into real images.
When training is complete, one would discard the training dataset and generate new samples 
using the decoder network alone. Hence, the burden of generative modeling rests entirely on the 
model parameters, and additional data cannot be incorporated without training.

This paper combines the advantages of  exemplar based and parametric 
methods using amortized variational inference, yielding a new generative model called Exemplar VAE.
It can be viewed as a variant of Variational Autoencoder (VAE)~\cite{kingma2013auto,rezende2014stochastic}
with a non-parametric Gaussian mixture (Parzen window) prior on latent codes.

To sample from the Exemplar VAE, one first draws a random exemplar from a training set, 
then stochastically transforms it into a latent code. 
A decoder than transforms the latent code into a new observation.
Replacing the conventional Gaussian prior into a non-parameteric Parzen window
improves the representation quality of VAEs as measured by kNN classification, presumably because
a Gaussian mixture prior with many components captures the manifold of images and their attributes better.
Exemplar VAE also improves density estimation on MNIST, Fashion MNIST, Omniglot, and CelebA, while
enabling controlled generation of images guided by exemplars.

We are inspired by recent work on generative models augmented with external memory 
(\eg~\cite{guu2018generating,li2019forest, tomczak2017vae,khandelwal2019generalization,bornschein2017variational}), 
but unlike most existing work, we do not rely on pre-specified distance metrics to define neighborhood structures.
Instead, we simultaneously learn an autoencoder, a latent space, and a  distance
metric by maximizing log-likelihood lower bounds.
We make critical technical contributions to make
Exemplar VAEs scalable to large datasets, and enhance their generalization.

The main contributions of this paper are summarized as follows: 
\vspace*{-.15cm}
\begin{enumerate}[topsep=0pt, partopsep=0pt, leftmargin=15pt, parsep=0pt, itemsep=1.75pt]
\item We introduce Exemplar VAE along with critical regularizers that combat overfitting;
\item We propose {\em retrieval augmented training (RAT)}, using approximate nearest neighbor search
in the latent space, to speed up training based on a novel log-likelihood lower bound;
\item Experimental results demonstrate that Exemplar VAEs consistently outperform VAEs with a Guassian 
prior or VampPrior~\cite{tomczak2017vae} on density estimation and representation learning;
\item We demonstrate  the effectiveness of generative data augmentation with Exemplar VAEs for 
supervised learning, reducing classification error of permutation invariant MNIST and Fashion MNIST 
significantly, from $1.17\%$ to $0.69\%$ and from 8.56\% to 8.16\% respectively.
\end{enumerate}





\vspace*{-.05cm}
\section{Exemplar based Generative Models}
\vspace*{-.1cm}

By way of background, an exemplar based generative model is defined in terms of 
a dataset of $N$ exemplars, $X \equiv \{\x_n\}_{n=1}^N$, and a parametric transition 
distribution, $\t_\theta(\x \mid \x')$, which stochastically transforms
an exemplar $\x'$ into a new observation $\x$. The log density of a data point
$\x$ under an exemplar based generative model $\{X, \t_\theta\}$ can be expressed as
\begin{equation}
\log p(\x \mid X, \theta) ~=~ \log \sum\nolimits_{n=1}^N\frac{1}{N} \t_\theta(\x \mid \x_n) ~,
\label{eq:exgen}
\end{equation}
where we assume the prior probability of selecting each exemplar is uniform.
Suitable transition distributions 
should place considerable probability mass on the reconstruction
of an exemplar from itself, \ie~$\t_\theta(\x \mid \x)$ should be large for all $\x$.
Further, an ideal transition distribution should be able to model the conditional dependencies
between different dimensions of $\x$ given $\x'$, since the dependence of $\x$ on $\x'$ is often
insufficient to make dimensions of $\x$ conditionally independent.

One can view the Parzen window or Kernel Density estimator~\citep{parzen1962estimation}, 
as a simple type of exemplar based generative model in which the transition 
distribution is defined in terms of a prespecified kernel function and its meta-parameters. 
With a Gaussian kernel, a Parzen window estimator takes the form
\begin{equation}
    \log p(\x \mid X, \sigma^2) ~=~ - \log C -\log N + \log \sum\nolimits_{n=1}^N \exp \frac{-\lVert \x - \x_n \rVert^2}{2\sigma^2}~,
\label{eq:parzen}
\end{equation}
where $\log C = d_{x} \log ( \sqrt{2 \pi} \sigma )$ is the log normalizing 
constant of an isotropic Gaussian in $d_{x}$ dimensions. 
The non-parametric nature of Parzen window estimators enables one to exploit extremely
large heterogeneous datasets of exemplars 
for density estimation.
That said, simple Parzen window estimation typically underperforms parametric density estimation, especially in high dimensional spaces, due to the inflexibility of typical transition distributions, 
\eg~when $\t(\x \mid \x') = \mathcal{N}(\x \mid \x', \sigma^2 I)$.

This work aims to adopt desirable properties of non-parametric exemplar based models to help 
scale parametric models to large heterogeneous datasets and representation learning. 
In effect, we learn a latent representation of the data for which a Parzen window estimator is an effective prior.

\vspace*{-.05cm}
\section{Exemplar Variational Autoencoders}
\label{sec:Methods}
\vspace*{-.1cm}

The generative process of an Exemplar VAE is summarized in three steps:
\vspace*{-.1cm}
\begin{enumerate}[topsep=0pt, partopsep=0pt, leftmargin=13pt, parsep=0pt, itemsep=2pt]
\item Sample $n \!\sim \!\mathrm{Uniform}(1, N)$ to obtain a random exemplar
  $\x_n$ from the training set, $X \equiv\{ \x_n\}_{n=1}^N$.
\item Sample $\z \!\sim\! r_\phi(\cdot \mid \x_n)$ using an exemplar based prior, $r_\phi$,
to transform an exemplar $\x_n$ into a distribution over latent codes, from which  $\z$ is drawn.
\item Sample $\x \!\sim\! p_\theta(\cdot \mid \z)$ using a
  decoder to transform $\z$ into a distribution over observations, from which $\x$ is drawn.
\end{enumerate}
\vspace*{-.05cm}
Accordingly, the Exemplar VAE can be interpreted as a variant of exemplar based generative models 
in \eqref{eq:exgen} with a parametric transition function defined in terms of a latent variable $\z$, \ie
\begin{equation}
\t_{\phi,\theta}(\x \mid \x') ~=~ \int_z r_\phi(\z \mid \x') \, p_\theta(\x \mid \z)\, d\z~.
\end{equation}
This model assumes that, conditioned on $\z$, an observation $\x$ is independent from an exemplar $\x'$. 
This conditional independence simplifies the formulation, enables efficient optimization, 
and encourages a useful latent representation.

\begin{figure}[t]
\begin{center}
\vspace*{-.4cm}
\includegraphics[width=.6\linewidth]{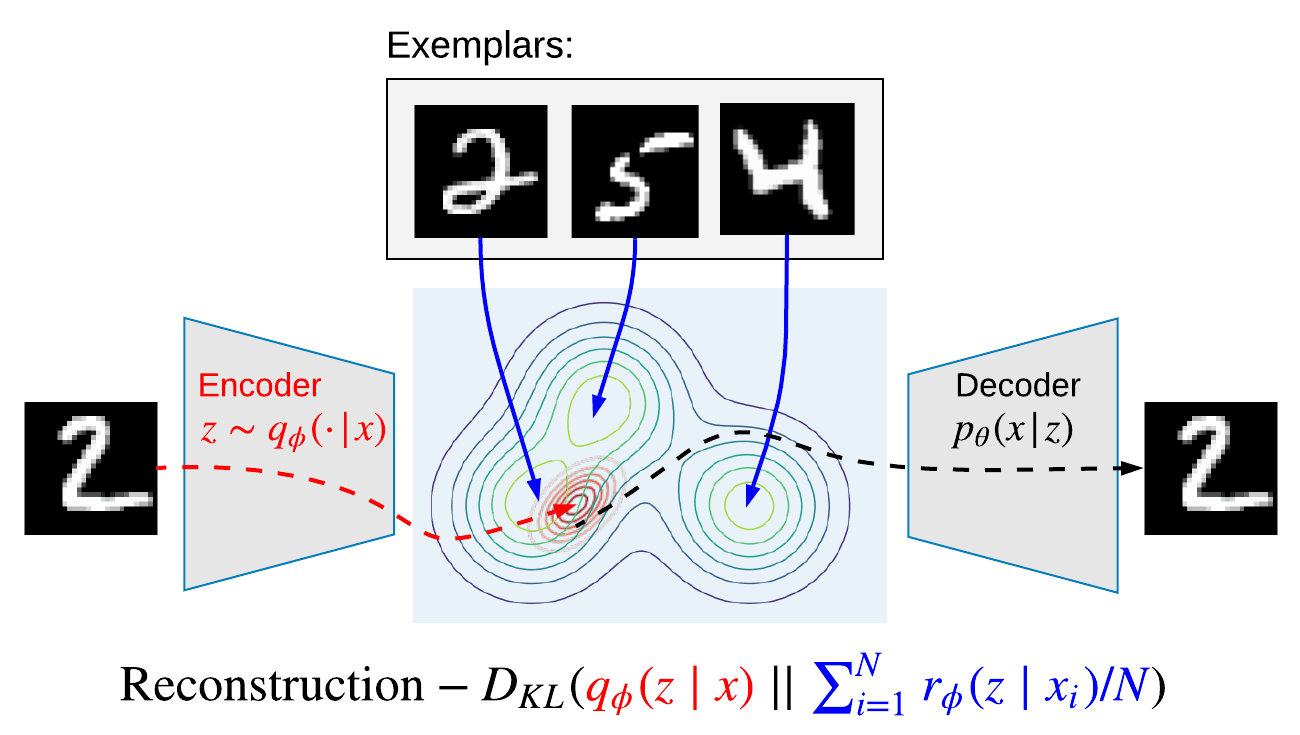} 
\vspace*{-.3cm}
\end{center}
\caption{\small Exemplar VAE is a type of VAE with a non-parametric mixture prior in the latent space.
Here, only $3$ exemplars are shown, but the set of exemplars often includes thousands of data points from the training dataset.
The objective function is similar to a standard VAE with the exception that the KL term measures the disparity between
the variational posterior $q_\phi(\z\!\mid\!\x)$ and a mixture of exemplar based priors $\sum_{n=1}^N r_\phi(\z \mid \x_n) / N$. 
\label{fig1}
}
\vspace*{-.3cm}
\end{figure}

By marginalizing over the exemplar index $n$ and the latent variable $\z$, one can derive an evidence lower bound (ELBO) \cite{blei2017variational,jordan1999introduction} on log marginal likelihood for a data point $\x$ as follows
(derivation in section F of supplementary materials):
\begin{align}
    \log p(\x; X, \theta, \phi)  &~=~
    \log \sum_{n=1}^N \frac{1}{N} \t_{\phi,\theta} (\x \mid \x_n) ~=~
    \log \sum_{n=1}^N \frac{1}{N}\int_z {r_\phi(\z \mid \x_n) \,p_\theta(\x \mid \z)}\, d\z\\
    &~\ge~ \underbrace{\mathop{\expected}_{q_{\phi}(\z \mid \x)} \!\!\!
    \log  p_{\theta}(\x\!\mid\!\z)}_{\mathrm{reconstruction}} ~ -\, \underbrace{\mathop{\expected}_{q_{\phi}(\z \mid \x)}
    \log \frac{q_\phi(\z \mid \x)}{\sum\nolimits_{n=1}^N r_\phi(\z \mid \x_n)/N}}_{\mathrm{KL~term}} \nonumber \\
    &~=~ O(\theta, \phi; \x, X).
    \label{eq:exVAE-ELBO}
\end{align}
We use \eqref{eq:exVAE-ELBO} as the Exemplar VAE objective to optimize parameters $\theta$ and $\phi$.
Note that $O(\theta, \phi; \x, X)$ is similar to the ELBO for a standard VAE,
the difference being the definition of the prior $p(\z)$ in the KL term.
The impact of exemplars on the learning objective can be summarized in the form of a mixture model
prior in the latent space, with one mixture component per exemplar, \ie
$p\left(\z \!\mid\! X \right) = \sum\nolimits_n\! r_\phi(\z \mid \x_{n})/N$.
Fig.\ \ref{fig1} illustrates the training procedure and objective function for Exemplar VAE.


A VAE with a Gaussian prior uses an encoder during training to define a variational bound~\cite{kingma2013auto}.  
Once training is finished, new observations are generated using the decoder network alone.
To sample from an Exemplar VAE, we need the decoder and access to a set of 
exemplars and the exemplar based prior $r_\phi$.
Importantly, given the non-parametric nature of Exemplar VAEs, one can train this model
with one set of exemplars and perform generation with another, potentially much larger set.

As depicted in \figref{fig1}, the Exemplar VAE employs two encoder networks, \ie ${q}_\phi(\z \! \mid
\!\x)$ as the variational posterior, and $r_\phi(\z \!\mid\! \x_{n})$ for mapping an exemplar $\x_n$ to
the latent space for the exemplar based prior. 
We adopt Gaussian distributions for both ${q}_\phi$ and $r_\phi$. 
To ensure that $T(\x\mid \x)$ is large, we share the means of ${q}_\phi$ and $r_\phi$.
This is also inspired by the VampPrior~\cite{tomczak2017vae} and discussions of
the aggregated variational posterior as a prior~\citep{makhzani2015adversarial,hoffman2016elbo}.
Accordingly, we define
\begin{align}
{q}_\phi(\z \mid \x) &~=~ \mathcal{N}(\z \mid \bmu_\phi(\x) \, , ~ \Lambda_\phi(\x)),\\
r_\phi(\z \mid \x_n) &~=~ \mathcal{N}(\z \mid \bmu_\phi(\x_n) \, ,~\sigma^2 I)~.
\label{eq:exprior}
\end{align}
The two encoders use the same parametric mean function $\bmu_\phi$, but 
they differ in their covariance functions. 
The variational posterior uses a data dependent diagonal covariance matrix ${\Lambda}_\phi$, 
while the exemplar based prior uses an isotropic Gaussian (per exemplar), with a shared, scalar parameter $\sigma^2$.
Accordingly, $\log p\left(\z \!\mid\! X\right)$, the log of the aggregated exemplar based prior is given by
\begin{equation}
    \log p\left(\z \!\mid\! X\right) ~=~
     - \log C' -\log N + \log\sum\nolimits_{j=1}^N \exp \frac{-\lVert \z - \bmu_\phi(\x_j) \rVert^2}{2\sigma^2} ~,
\label{eq:latent-kde}
\end{equation}
where $\log C' = d_{z} \log ( \sqrt{2 \pi} \sigma )$. 
Recall the definition of Parzen window estimator with a Gaussian kernel in \eqref{eq:parzen}, 
and note the similarity between \eqref{eq:parzen} and \eqref{eq:latent-kde}. 
The Exemplar VAE's Gaussian mixture prior is a Parzen window estimate in the latent space,
hence the Exemplar VAE can be interpreted as a {\em deep} variant of Parzen window estimation.

The primary reason to adopt a shared $\sigma^2$ across exemplars in \eqref{eq:exprior} is computational efficiency.
Having a shared $\sigma^2$ enables parallel computation of all pairwise distances between a minibatch 
of latent codes $\{\z_b\}_{b=1}^B$ and Gaussian means $\{\bmu_\phi(\x_j)\}_{j=1}^N$ using a single matrix product.
It also enables the use of existing approximate nearest neighbor search 
methods for Euclidean distance~(\eg~\cite{muja2014scalable}) to speed up Exemplar VAE training, as described next.

\vspace*{-0.05cm}
\subsection{Retrieval Augmented Training (RAT) for Efficient Optimization}
\label{sec:NNsearch}
\vspace*{-0.1cm}

The computational cost of training an Exemplar VAE can become a burden as the number of exemplars increases. 
This can be mitigated with fast, approximate nearest neighbor search in the latent space to 
find a subset of exemplars that exert the maximum influence on the generation of each data point.
Interesting, as shown below, the use of approximate nearest neighbor for training Exemplar VAEs is mathematically
justified based on a lower bound on the log marginal likelihood.

The most costly step in training an Exemplar VAE is in the computation of $\log p\left(\z \!\mid\! X\right)$ 
in \eqref{eq:latent-kde} given a large dataset of exemplars $X$, where $\z \sim q_{\phi}(\z \mid \x)$ 
is drawn from the variational posterior of $\x$. 
The rest of the computation, to estimate the reconstruction error and the entropy of the 
variational posterior, is the same as a standard VAE.
To speed up the computation of $\log p\left(\z \!\mid\! X\right)$, 
we evaluate 
$\z$ against $K \ll N$ exemplars that exert the maximal influence on $\z$, and ignore the rest. 
This is a reasonable approximation in high dimensional spaces where only the nearest Gaussian
means matter in a Gaussian mixture model.
Let $\mathrm{kNN}(\z) \equiv \{\pi_k\}_{k=1}^K$ denote the set of $K$ exemplar indices with approximately 
largest $r_\phi(\z \!\mid\! \x_{\pi_k})$, or equivalently, the smallest $\lVert \z - \bmu_\phi(\x_{\pi_k}) \rVert^2$
for the model in \eqref{eq:exprior}. Since probability densities are non-negative and $\log$ is monotonically 
increasing, it follows that
\begin{equation}
    \log p\big( \z \!\mid\! X \big) 
    ~=~ -\log {N} + \log \sum_{j=1}^N {r_\phi(\z \!\mid\! \x_{j})} 
    ~\ge~ -\log {N} + \log \!\!\sum_{k \in \mathrm{kNN}(\z)} \!\!\! {r_\phi(\z \!\mid\! \x_{\pi_k})} ~
\label{eq:knn-bound}
\end{equation}
As such, approximating the exemplar prior with approximate kNN is a lower bound on \eqref{eq:latent-kde} and \eqref{eq:exVAE-ELBO}.

To avoid re-calculating $\{\bmu_\phi(\x_j)\}_{j=1}^N$ for each gradient update, we store a 
cache table of most recent latent means for each exemplar. Such cached latent means are used for
approximate nearest neighbor search to find $\mathrm{kNN}(\z)$. Once approximate kNN indices are found, the latent means,
$\{\bmu_\phi(\x_{\pi_k})\}_{k \in \mathrm{kNN}(\z)}$, are re-calculated to ensure that the bound in \eqref{eq:knn-bound} is valid.
The cache is updated whenever a new latent mean of a training point is available,
\ie~we update the cache table for any point covered by the training minibatch or the kNN exemplar sets.
Section C in the supplementary materials summaries the Retrieval Augmented Training (RAT) procedure.



\vspace*{-0.05cm}
\subsection{Regularizing the Exemplar based Prior}
\label{sec:reg}
\vspace*{-0.1cm}

Training an Exemplar VAE by simply maximizing $O(\theta, \phi; \x, X)$ in \eqref{eq:exVAE-ELBO},
averaged over training data points $\x$, often yields massive overfitting. 
This is not surprising, since a flexible transition distribution can put all its probability 
mass on the reconstruction of each exemplar, \ie~$p(\x \mid \x)$, yielding high log-likelihood 
on training data but poor generalization. 
Prior work~\cite{bornschein2017variational,tomczak2017vae} also observed such overfitting, 
but no remedies have been provided.
To mitigate overfitting we propose two simple but effective {\em regularization} strategies:
\begin{enumerate}[topsep=0pt, partopsep=0pt, leftmargin=10pt, parsep=0pt, itemsep=3pt]
    \item {\bf Leave-one-out during training.} The generation of a given data point is expressed 
    in terms of dependence on all exemplars except that point itself. The non-parametric nature of the generative 
    model enables easy adoption of such a leave-one-out (LOO) objective during training,
    to optimize
    \begin{equation}
    O_1(\phi,\theta; X) ~=~ 
    \sum\nolimits_{i=1}^N \log \sum\nolimits_{n=1}^N \frac{\one{i \neq n}}{N\!-\!1} \t_{\phi,\theta}(\x_i \mid \x_n)~,
    \end{equation}
    where $\one{i \neq n} \in \{0, 1\}$ is an indicator function, taking the value of $1$ if and only if $i \neq n$.

    \item {\bf Exemplar subsampling.} 
    Beyond LOO, we observe that explaining a training point using a subset of 
    the remaining training exemplars improves generalization.
    To that end, we use a hyper-parameter $M$ to define the exemplar subset size for the generative model.
    To generate $\x_i$ we draw $M$ indices $\pi \equiv \{\pi_m\}_{m=1}^M$ 
    uniformly at random from subsets of $\{1, \ldots, N\} \setminus \{ i\}$. 
    Let $\pi \sim \Pi^{N,i}_{M}$ denote this sampling procedure with 
    ($N\!-\!1$ choose $M$) possible subsets.      
    This results in the objective function\vspace*{-.1cm}
    \begin{equation}
    O_2(\phi,\theta; X) ~=~ 
    \sum\nolimits_{i=1}^N \mathop{\expected~~~~~~~~~}_{\pi\sim~\Pi^{N,i}_{M}} \log \sum\nolimits_{m=1}^M 
    \frac{1}{M} \t_{\phi,\theta}(\x_i \mid \x_{\pi_m}) ~.
    \label{eq:obj2}
    \end{equation}
    By moving $\expected_\pi$ inside the log in \eqref{eq:obj2} we recover $O_1$; \ie $O_2$ is a lower bound on $O_1$, via Jensen's inequality.
    Interestingly, we find $O_2$ often yields better generalization than $O_1$.
\end{enumerate}

Once training is finished, all $N$ training exemplars are used to
explain the generation of the validation or test sets using \eqref{eq:exgen},
for which the two regularizers discussed above 
are not used.
Even though cross validation is commonly used for parameter tuning and model selection, in \eqref{eq:obj2}
cross validation is used as a training objective directly, suggestive of a {\em meta-learning} perspective.
The non-parameteric nature of the exemplar based prior enables the use of the regularization 
techniques above, but this would not be straightforward for training parametric generative models. 


{\bf Learning objective.}~To complete the definition of the learning objective for an Exemplar VAE, we 
combine RAT and exemplar sub-sampling to obtain the final Exemplar VAE objective:
\begin{equation}
    O_3(\theta, \phi; X) =
    \sum_{i=1}^N \mathop{\expected}_{q_{\phi}(\z \mid \x_i)}\!\left[
    \log \frac{p_{\theta}(\x_i\!\mid\!\z)}{q_\phi(\z \!\mid\! \x_i)} +
    \!\!\mathop{\expected}_{\Pi^{N,i}_{M}(\pi)}
    \!\!\! \log \sum_{m=1}^M \frac{\one{\pi_m \in \mathrm{kNN}(\z)}}{(\sqrt{2 \pi} \sigma)^{d_z}} \exp \frac{-\lVert \z - \bmu_\phi(\x_{\pi_m}) \rVert^2}{2\sigma^2}
    \right],
\label{eq:exVAE}
\end{equation}
where, for brevity, the additive constant $-\log M$ has been dropped. 
We use the  reparametrization trick to back propagate through $\expected \,q_{\phi}(\z \mid \x_i)$. 
For small datasets and fully connected architectures we do not use RAT, but for convolutional 
models and large datasets the use of RAT is essential.

\vspace*{-0.1cm}
\section{Related Work}
\vspace*{-0.1cm}


Variational Autoencoders (VAEs)~\cite{kingma2013auto,rezende2014stochastic} are versatile,
latent variable generative models, used for non-linear dimensionality 
reduction~\cite{gregor2016towards}, generating discrete data~\cite{bowman2015generating}, 
and learning disentangled representations \cite{higgins2016beta, chen2018isolating}, while
providing a tractable lower bound on log marginal likelihood.
Improved variants of the VAE are based on modifications to the VAE objective~\cite{burda2015importance}, 
more flexible variational familieis~\cite{kingma2016improved,rezende2015variational}, 
and more powerful decoders~\cite{chen2016variational,gulrajani2016pixelvae}.
More powerful latent priors \cite{tomczak2017vae,bauer2018resampled,dai2018diagnosing,lawson2019energy} 
can improve the effectiveness of VAEs for density estimation, as suggested by \cite{hoffman2016elbo}, 
and motivated by the observed gap between the prior and aggregated posterior~(\eg~\cite{makhzani2015adversarial}). 
More powerful priors may help avoid posterior collapse in VAEs with autoregressive decoders~\cite{bowman2015generating}.
Unlike most existing work, Exemplar VAE assumes little  about the structure of the latent space, 
using a non-parameteric  prior.


VAEs with a VampPrior \cite{tomczak2017vae} optimize a set of pseudo-inputs together 
with the encoder network to obtain a Gaussian mixture approximation to the aggregate posterior.
They argue that computing the exact aggregated posterior, while desirable, is
expensive and suffers from overfitting, hence they restrict the number 
of pseudo-inputs to be much smaller than the training set.
Exemplar VAE enjoys the use of all training points, but without a large increase
in the the number of model parameters, while avoiding overfitting through simple regularization techniques.
Training cost is reduced through RAT using approximate kNN search during training.

Exemplar VAE also extends naturally to large high dimensional datasets, and
to discrete data, without requiring additional pseduo-input parameters. 
VampPrior and Exemplar VAE are  similar in their reuse of the encoder network 
and a mixture prior over the latent space. However, the encoder for the  
Exemplar VAE prior has a simplified covariance, which is useful for efficient learning.
Importantly, we show that Exemplar VAEs can learn better unsupervised representations of images and
perform generative data augmentation to improve supervised learning.


Memory augmented networks with attention can enhance generative models~\cite{li2016learning}.
Hard attention has been used in VAEs~\cite{bornschein2017variational}
to generate images conditioned on memory items, with learnable and fixed memories.  
One can view Exemplar VAE as a VAE with external memory.  
One crucial difference between Exemplar VAE and \cite{bornschein2017variational} 
is in the conditional dependencies assumed in the Exemplar VAE, which disentangles the prior and reconstruction terms,
and enables amortized computation per minibatch.
In \cite{bornschein2017variational} discrete indices are optimized which creates challenges for gradient estimation,
and they need to maintain a normalized categorical distribution over a potentially massive set of indices.
By contrast, we use approximate kNN search in latent space to model hard attention, without requiring a normalized categorical distribution or high variance gradient estimates, and we mitigate overfitting using regularization.

Associative Compression Networsk \cite{graves2018associative} learn an ordering over a dataset to obtain better compression rates through VAEs. That work is similar to ours in defining the prior based on training data samples and the use of kNN in the latent space during training. However, their model with a conditional prior is not comparable with order agnostic VAEs. On the other hand, Exemplar VAE has an unconditional prior where, after training, defining an ordering is feasible and achieves the same goal

\vspace*{-.1cm}
\section{Experiments}
\vspace*{-.1cm}

\textbf{Experimental setup.}~
We evaluate Exemplar VAE on density estimation, representation learning, and data augmentation.
We use four datasets, namely, MNIST, Fashion-MNIST, Omniglot, and CelebA,
and we consider four different architectures for gray-scale image data,  namely, a VAE with MLP for encoder and decoder with two hidden layers 
(300 units each), a HVAE with similar architecture but two stochastic layers, ConvHVAE with two stochastic layers and convolutional encoder and decoder, and PixelSNAIL \cite{chen2018pixelsnail} with two stochastic layers and an auto-regressive PixelSNAIL shared between encoder and decoder. For CelebA we used a convolutional architecture based on \cite{ghosh2019variational}.
We use gradient normalized Adam  \cite{kingma2014adam,yu2017block} with learning 
rate 5e-4  and linear KL annealing for 100 epochs. 
See the supplementary material  for details.



\textbf{Evaluation.} For density estimation we use Importance Weighted Autoencoders 
(IWAE)~\cite{burda2015importance} with 5000 samples, using the entire 
training set as exemplars, without regularization or kNN acceleration. 
This makes the evaluation time consuming, but generating an unbiased sample from the Exemplar
VAE is efficient.  Our preliminary experiments suggest that using kNN 
for evaluation is feasible. 


\vspace{-.1cm}
\subsection{Ablation Study}
\vspace{-.1cm}

First, we evaluate the effectiveness of the regularization techniques
proposed (\figref{figure:1}), \ie~leave-one-out and exemplar
subsampling, for enhancing generalization.

\textbf{Leave-one-out (LOO).}~We train an Exemplar VAE with a full aggregated exemplar based prior without RAT
with and without LOO. \figref{figure:1} plots the ELBO computed on training and validation sets,
demonstrating the surprising effectiveness of LOO in regularization. 
\tabref{table-log-like} gives test log-likelihood IWAE bounds for Exemplar VAE 
on MNIST and Omniglot with and without LOO. 

\begin{table*}[h]
\small
\begin{minipage}[h!][][t]{0.48\textwidth}
\centering
\includegraphics[width=0.85\textwidth]{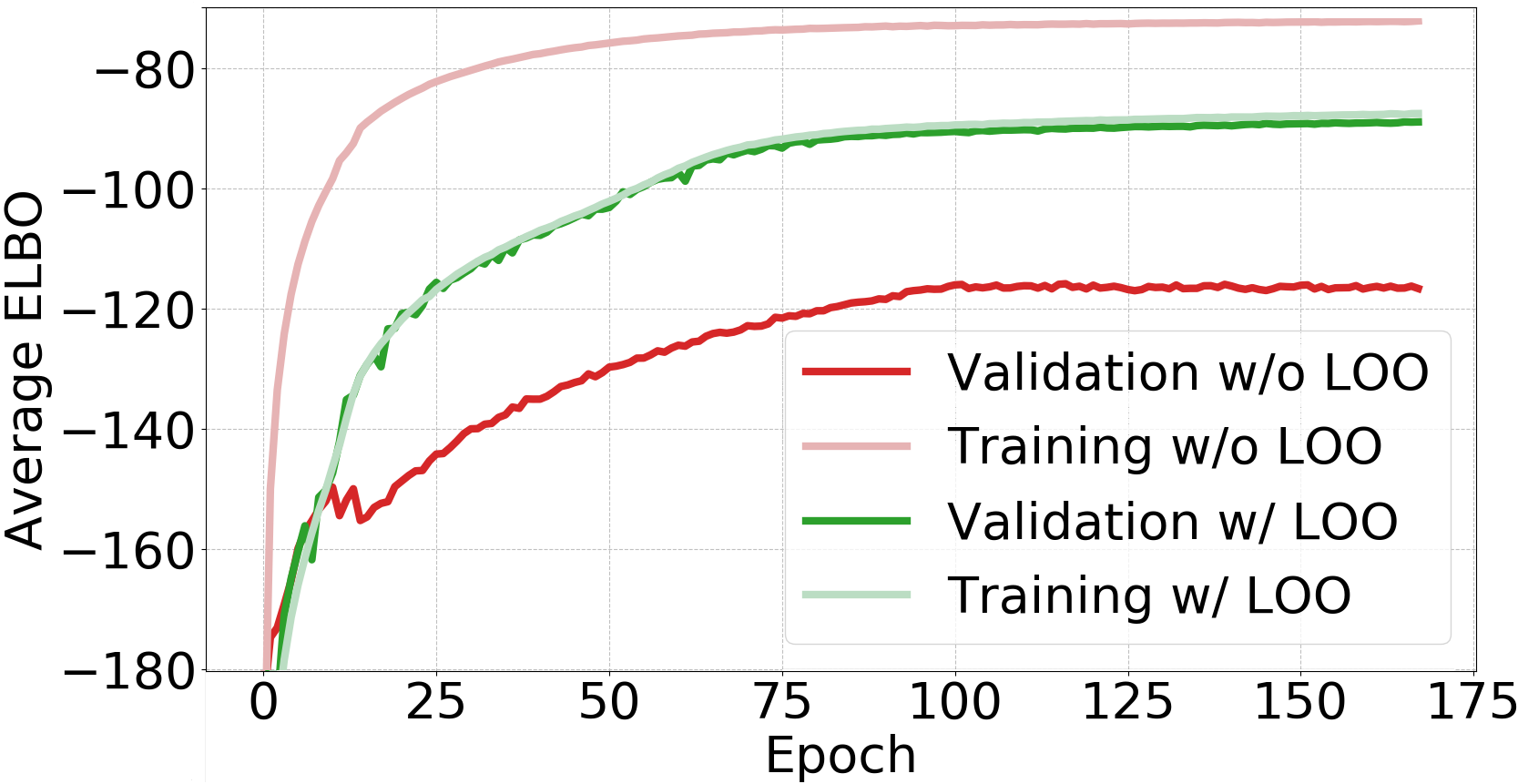}
\vspace{-.1cm}
\captionof{figure}{\small Training and validation ELBO on Dynamic MNIST for Exemplar VAE with and without LOO. }
\label{figure:1}
\end{minipage}
\hfill
\begin{minipage}[h!][][t]{0.45\textwidth}
\small
\centering
\begin{tabular}{c c c} 
 \toprule
 & \multicolumn{2}{c}{Exemplar VAE}\\
 {Dataset} & w/ LOO & w/o LOO\\[0.5ex] 
 \midrule
 MNIST & $-82.35$ & $-101.33$\\ 
 Omniglot & $-105.80$ & $-139.12$\\
 \bottomrule
\end{tabular}
\vspace*{-0.01cm}
 \caption{\small Log likelihood lower bounds on the test set (nats) for Exemplar VAE 
 with and without leave-one-out (LOO).
 \label{table-log-like}
 }
\end{minipage}
\vspace{-.1cm}

\end{table*}

\textbf{Exemplar subsampling.}~As explained in \secref{sec:reg},
the Exemplar VAE uses a hyper-parameter $M$ to define the number of exemplars used
for estimating the prior.  Here, we report the Exemplar VAE's density estimates
as a function of $M$ divided by the number of training data points $N$.
We consider $M/N \in \{1.0, 0.5, 0.2, 0.1\}$. 
All  models use LOO, and $M/N=1$ reflects $M=N-1$.
\tabref{table:ablation-subsampling} presents results for MNIST and Omniglot.
In all of the following experiments we adopt $M/N = 0.5$.

\begin{table*}[h!]
\vspace*{-0.1cm}
\small
\centering
 \begin{tabular}{ccccc} 
 \toprule
 \diagbox[height=.55cm]{Dataset}{$M/N$} & 1 & 0.5 & 0.2 & 0.1 \\
 \midrule
 {MNIST} & {$-82.35$} & {$\mathbf{-82.09}$} & {$-82.12$} & {$-82.20$}\\ 
 {Omniglot} & {$-105.80$} & {$-105.22$}& {$\mathbf{-104.95}$} & {$-105.42 $}\\ 
 \bottomrule
\end{tabular}
\vspace*{-.01cm}
\caption{\small Test log likelihood lower bounds (nats) for Exemplar VAE versus fraction of exemplar subsampling.
\label{table:ablation-subsampling}
}
\vspace*{-.1cm}
\end{table*}

\begin{table*}[t]
\vspace*{-0.2cm}
\small
\begin{center}
\begin{tabular}{rcccc}
\toprule
Method && Dynamic MNIST & Fashion MNIST & Omniglot \\ 
\midrule
VAE w/ Gaussian prior  &&  $-84.45$ {\footnotesize $\pm 0.12$}& $-228.70$ {\footnotesize $\pm 0.15$} & $-108.34$ {\footnotesize $\pm 0.06$} \\
VAE w/ VampPrior       &&  $-82.43$ {\footnotesize $\pm0.06$} & $-227.35$ {\footnotesize $\pm 0.05$} & $-106.78$ {\footnotesize $\pm 0.21$} \\
Exemplar VAE           &&  {\boldmath $-82.09$} {\footnotesize $\pm 0.18$} & {\boldmath $-226.75$} {\footnotesize $\pm 0.07$} & {\boldmath $-105.22$} {\footnotesize $\pm 0.18$}  \\
\midrule
HVAE w/ Gaussian prior && $-82.39$ {\footnotesize $\pm  0.11$} & $227.37$ {\footnotesize $\pm 0.1$} & $-104.92$ {\footnotesize $\pm 0.08$} \\
HVAE w/ VampPrior  &&  $-81.56$ {\footnotesize $\pm 0.09$} & $-226.72$ {\footnotesize $\pm 0.08$} & $-103.30$ {\footnotesize $\pm 0.43$}  \\
Exemplar HVAE && {\boldmath $-81.22$} {\footnotesize $\pm 0.05$} & {\boldmath $-226.53$} {\footnotesize $\pm 0.09$} &  {\boldmath $-102.25$} {\footnotesize $\pm 0.43$} \\
\midrule
ConvHVAE w/ Gaussian prior&& $-80.52$ {\footnotesize $\pm 0.28$} & $-225.38$ {\footnotesize $\pm 0.08$} & $-98.12$ {\footnotesize $\pm0.17$}  \\
ConvHVAE w/ Lars && $-80.30$ & $-225.92$  & $-97.08$ \\
ConvHVAE w/ SNIS && $-79.91$ {\footnotesize$\pm 0.05$} & $-225.35$ {\footnotesize$\pm 0.07$} & N/A  \\
ConvHVAE w/ VampPrior && $-79.67$ {\footnotesize $\pm 0.09$} & $-224.67$ {\footnotesize $\pm 0.03$ } & $-97.30$ {\footnotesize $\pm 0.07$ }  \\
Exemplar ConvHVAE && {\boldmath $-79.58$} {\footnotesize $\pm 0.07$} & {\boldmath$-224.63$} {\footnotesize $\pm 0.06$} & {\boldmath $-96.38$} {\footnotesize $\pm 0.24$} \\
\midrule
PixelSNAIL w/ Gaussian Prior && $-78.20$ {\footnotesize $\pm 0.02$} & $-223.68$ {\footnotesize $\pm 0.03$ } & $-89.59$ {\footnotesize $\pm 0.07$ }  \\
PixelSNAIL w/ VampPrior && {\boldmath $-77.90$} {\footnotesize $\pm 0.02$} & $-223.45$ {\footnotesize $\pm 0.02$ } & $-89.50$ {\footnotesize $\pm 0.13$ }  \\
Exemplar PixelSNAIL && $-77.95$ {\footnotesize $\pm 0.01$} & {\boldmath$-223.26$} {\footnotesize $\pm 0.01$} & {\boldmath $-89.28$} {\footnotesize $\pm 0.12$} \\
\bottomrule
\end{tabular}
\end{center}
\vspace*{-0.1cm}
\caption{\small Density estimation on dynamic MNIST, Fashion MNIST,
and Omniglot for different methods and architectures, all with 40-D latent spaces.
Log likelihood lower bounds (nats), estimated with IWAE with 5000 samples, are averaged over 5 training runs.
For LARS~\cite{bauer2018resampled} and SNIS~\cite{lawson2019energy}, the IWAE used 1000 samples; 
their architectures and training procedures are also somewhat different.
\label{tab:loglikelihood}
}
\vspace*{-0.01cm}
\end{table*}

\begin{figure*}[t]
\vspace*{-0.01cm}
\small
\begin{center}
\begin{tabular}{@{}
c@{\hspace{0.01\linewidth}}c@{\hspace{0.01\linewidth}}c@{\hspace{0.018\linewidth}}
c@{\hspace{0.01\linewidth}}c@{\hspace{0.01\linewidth}}c@{\hspace{0.018\linewidth}}
c@{\hspace{0.01\linewidth}}c@{\hspace{0.01\linewidth}}c@{}
}

\includegraphics[width=0.1\linewidth]{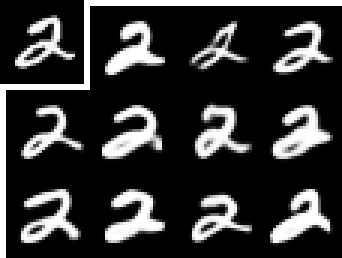} &
\includegraphics[width=0.1\linewidth]{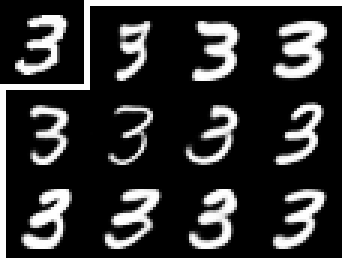} &
\includegraphics[width=0.1\linewidth]{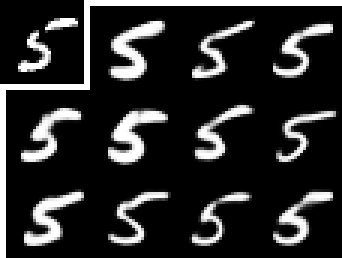} &
\includegraphics[width=0.1\linewidth]{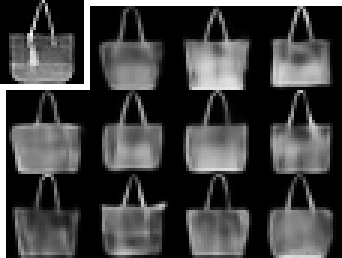} &
\includegraphics[width=0.1\linewidth]{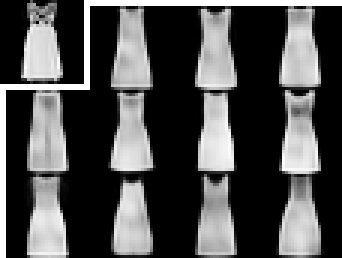} &
\includegraphics[width=0.1\linewidth]{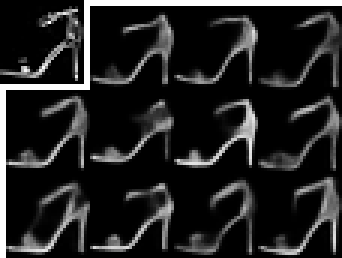} &
\includegraphics[width=0.1\linewidth]{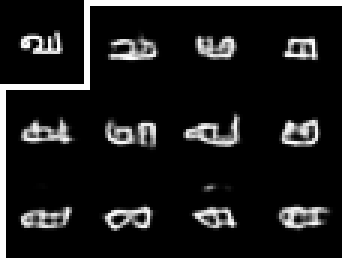} &
\includegraphics[width=0.1\linewidth]{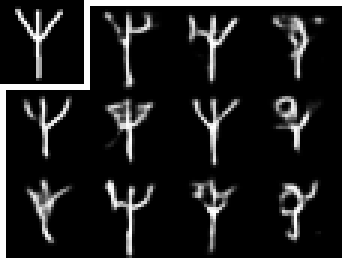} &
\includegraphics[width=0.1\linewidth]{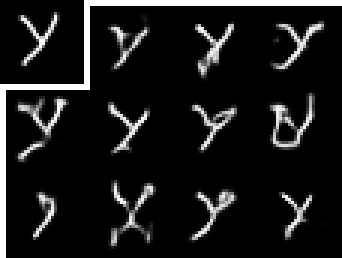} 
\\
\includegraphics[width=0.1\linewidth]{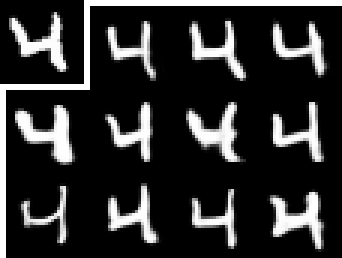} &
\includegraphics[width=0.1\linewidth]{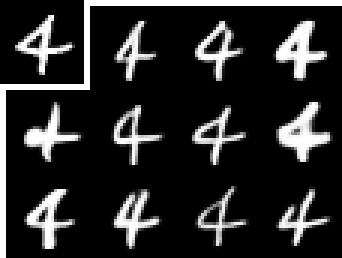} &
\includegraphics[width=0.1\linewidth]{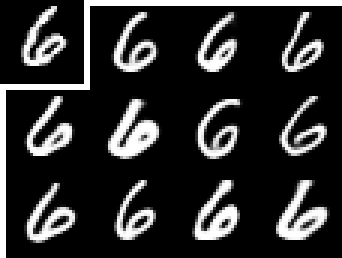} &
\includegraphics[width=0.1\linewidth]{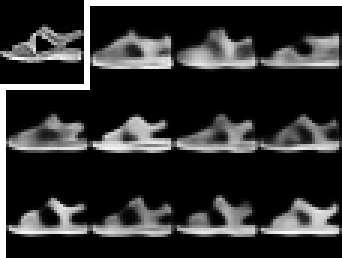} &
\includegraphics[width=0.1\linewidth]{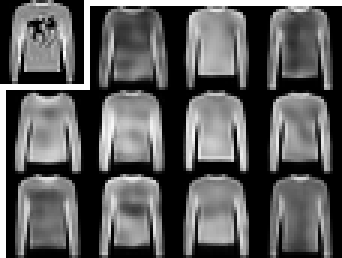} &
\includegraphics[width=0.1\linewidth]{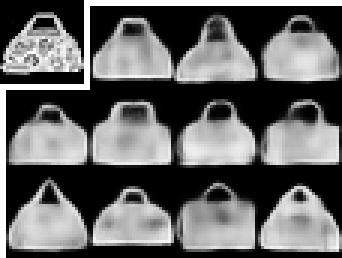} &
\includegraphics[width=0.1\linewidth]{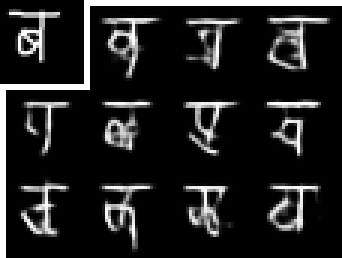} &
\includegraphics[width=0.1\linewidth]{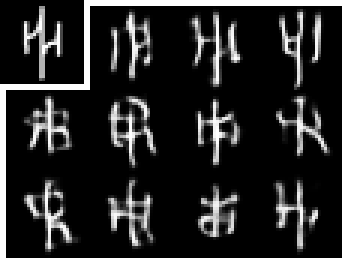} &
\includegraphics[width=0.1\linewidth]{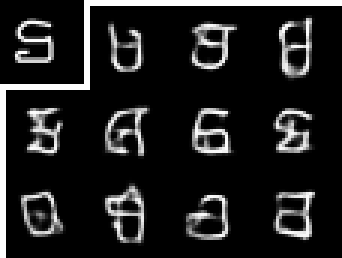}
\\

\multicolumn{3}{c}{\small MNIST} & \multicolumn{3}{c}{\small Fashion MNIST} & \multicolumn{3}{c}{\small Omniglot}
\\
\end{tabular}
\end{center}
\vspace*{-0.1cm}

\small
\begin{center}
\begin{tabular}{@{}
c@{\hspace{0.01\linewidth}}
c@{\hspace{0.01\linewidth}}
c@{\hspace{0.01\linewidth}}
c@{\hspace{0.01\linewidth}}
c@{\hspace{0.01\linewidth}}
}
\includegraphics[width=0.19\linewidth]{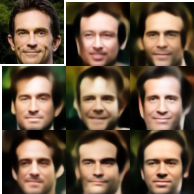}  
& \includegraphics[width=0.19\linewidth]{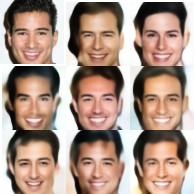} 
& \includegraphics[width=0.19\linewidth]{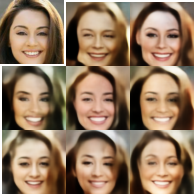} 
& \includegraphics[width=0.19\linewidth]{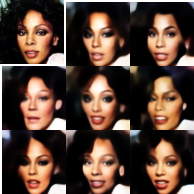} 
& \includegraphics[width=0.19\linewidth]{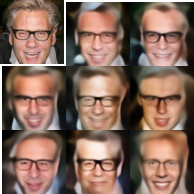} 

\\
\multicolumn{5}{c}{\small CelebA}\\
\end{tabular}
\end{center}
\vspace*{-0.1cm}

\caption{\small Given a source exemplar on the top left of each plate, 
Exemplar VAE samples are generated, showing a significant diversity while preserving 
properties of the source exemplar.
\label{figure:sampling}
}
\label{figure:celebA-generation}
\vspace*{-.1cm}
\end{figure*}

\comment{
\begin{figure*}[t]
\small
\begin{center}
\begin{tabular}{@{}c@{\hspace{0.04\linewidth}}c@{\hspace{0.04\linewidth}}c@{}}
\includegraphics[width=0.3\linewidth]{images/mnist_generation.png} &
\includegraphics[width=0.3\linewidth]{images/fashion_mnist_generation.png} &
\includegraphics[width=0.32\linewidth]{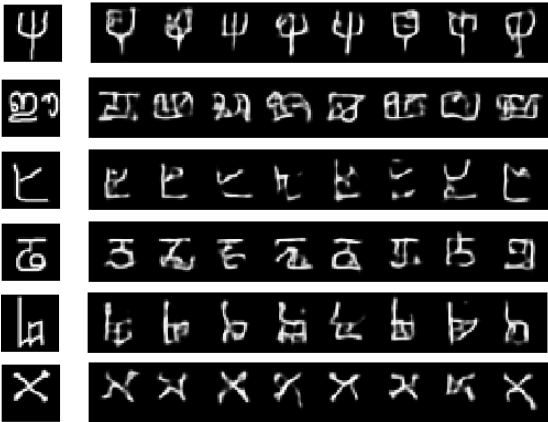}\\
MNIST & Fashion MNIST & Omniglot
\end{tabular}
\end{center}
\vspace*{-0.1cm}
\caption{\small Samples generated by the Exemplar VAE.
Given the input exemplar on the left, $8$ conditional model samples
are shown on the right.  Note that Exemplar VAE is able to preserve
the style and identity of the input exemplar, yet generates novel and
diverse samples.
\label{figure:mnist-generation}
}
\end{figure*}
}

\begin{figure}[h]
\small
\vspace*{-0.1cm}
\begin{center}
\begin{tabular}{@{}c@{\hspace{.1cm}}c@{}}
     \includegraphics[width=0.49 \linewidth]{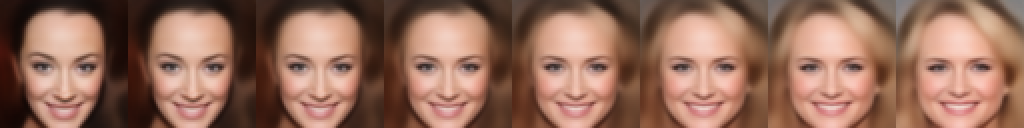} & 
     \includegraphics[width=0.49 \linewidth]{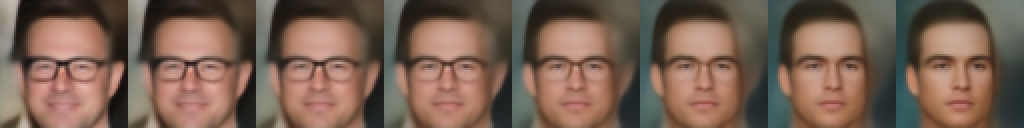} \\
     \includegraphics[width=0.49 \linewidth]{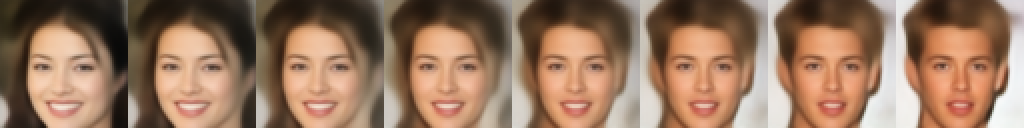}
     & \includegraphics[width=0.49 \linewidth]{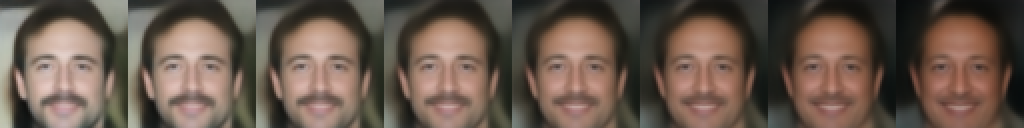}
\end{tabular}
\end{center}
\vspace*{-0.3cm}
\caption{\small Interpolation between samples from the CelebA dataset. }
\label{fig:celebA-interpolation}
\vspace*{-0.1cm}
\end{figure}

\vspace*{-.1cm}
\subsection{Density Estimation}
\vspace*{-.1cm}

For each architecture, we compare to a Gaussian prior and a VampPrior, which 
represent the state-of-the-art among VAEs with a factored variational posterior.
For training VAE and HVAE we did not utilize RAT, but for convolutional architectures 
we used RAT with 10NN search (see Sec.\ \ref{sec:NNsearch}). Note that the number of nearest neighbors are selected based on computational budget; we believe larger values work better.
Table \ref{tab:loglikelihood} shows that Exemplar VAEs outperform other models in all cases except one.
Improvement on Omniglot is greater than on other datasets, which may be due to its significant diversity. 
One can attempt to increase the number of pseudo-inputs in VampPrior, but this leads to overfitting.
As such, we posit that Exemplar VAEs have the potential to more easily  scale to large, diverse datasets. 
Note that training an Exemplar ConHVAE with approximate 10NN search is as efficient as training a ConHVAE with a VampPrior. 
Also, note that VampPrior \cite{tomczak2017vae} showed that a mixture of variational posteriors outperforms a Gaussian mixture prior, and hence we do not directly compare to that baseline.

Fig.\ \ref{figure:sampling} shows samples generated from an Exemplar ConvVAE, for which the corresponding exemplars are shown in the top left corner of each plate. 
These samples highlight the power of Exemplar VAE in maintaining the content of the source exemplar while adding diversity. 
For MNIST the changes are subtle, 
but for Fashion MNIST and Omniglot samples show more pronounced variation in style, possibly because those datasets are more diverse. 

To assess the scalability of Exemplar VAEs to larger datasets, 
we train this model on $64\!\times \!64$ CelebA images \cite{liu2015faceattributes}.
Pixel values are modeled using a discretized logistic distribution \cite{kingma2016improved,salimans2017pixelcnn++}.
Exemplar VAE samples (\figref{figure:sampling}) are high quality with good diversity.
Interpolation in the latent space is also effective (\figref{fig:celebA-interpolation}).
More details and quantitative evaluations
are provided in the supplementary materials.



\vspace*{-.1cm}
\subsection{Representation Learning}
\vspace*{-.1cm}

\begin{figure}[t]
\vspace*{-0.1cm}

\begin{minipage}[h!][][t]{0.44\textwidth}
\begin{center}
\small
  \begin{tabular}{@{}c@{\hspace{0.015\linewidth}}c@{}}
    \includegraphics[width=2.6cm]{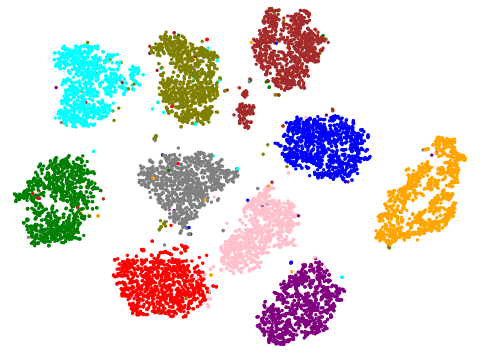} & \includegraphics[width=2.6cm]{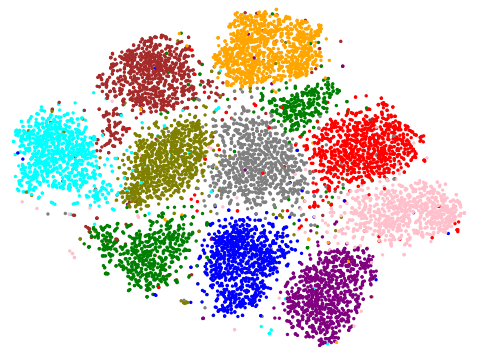} \\
    \scriptsize{Exemplar VAE on MNIST} & \scriptsize{VAE on MNIST} \\
  \end{tabular}
\end{center}
\vspace*{-0.1cm}
\caption{\small t-SNE visualization of learned latent representations for test points, colored by labels.
\label{fig:tSNE-MNIST}}
\end{minipage}
\hfill
\begin{minipage}[h!][][t]{0.52\textwidth}
\begin{center}
\small
\begin{tabular}{@{\hspace{.1cm}}l @{\hspace{.3cm}}l@{\hspace{.3cm}} l@{\hspace{.1cm}}} 
 \toprule
 Method & MNIST & Fashion MNIST\\[0.5ex] 
 \midrule
 VAE w/ Gaussian Prior & $2.41$ \footnotesize{$\pm0.27$} &  $15.90$ \footnotesize{$\pm0.34$} \\
 VAE w/ VampPrior & $1.42$ \footnotesize{$\pm0.02$}& $12.74$ \footnotesize{$\pm0.18$} \\ 
 Exemplar VAE & {\boldmath $1.13$} \footnotesize{$\pm0.06$} & {\boldmath $12.56$} \footnotesize{$\pm0.08$} \\ [1ex] 
 \bottomrule
\end{tabular}
\end{center}
\vspace*{-0.15cm}
\captionof{table}{\small kNN classification error (\%) on 40-D unsupervised representations.
\label{table:test_knn}
}
\end{minipage}
\vspace*{-0.1cm}
\end{figure}

We next explore the structure of the latent representation for Exemplar VAE. 
Fig.\ \ref{fig:tSNE-MNIST} shows a t-SNE visualization of the latent 
representations of MNIST test data for the Exemaplar VAE and for VAE with a Gaussian prior.
Test points are colored by their digit label.
No labels were used during training.
The Exemplar VAE representation appears more meaningful, 
with tighter clusters than VAE.
We also use k-nearest neighbor (kNN) classification performance as a proxy 
for the representation quality.
As is clear from \tabref{table:test_knn}, 
Exemplar VAE consistently outperforms other approaches. 
Results on Omniglot are not reported since
the low resolution variant of this dataset does not include class labels.
We also counted the number of active dimension in the latent to measure posterior collapse. 
Section D of supplementary materials shows the superior behavior of Exemplar VAE.

\comment{
\subsection{Generative Models Evaluation}
Exemplar VAE also benefits from some properties of non-parametric models. For instance, After fitting an Exemplar VAE to a particular dataset we should not necessarily use the original training data for evaluation. We can take samples from any generative model (GANs, Energy Based Models, VAEs) and use those samples as a exemplar for Exemplar VAE. If a generative model captures the diversity and properties of the training dataset well, we believe the evaluation of test-set probability under exemplar VAE based on samples from that generative should not be that different from Exemplar VAE with samples from original training data. Further more we can rank different generative model by this evaluation.\par

We evaluated three VAE variants trained in previous section. We sampled 50k from each generative model and used them as an exemplar for

\begin{table}[h!]
\centering
\small
 \begin{tabular}{@{}c c c c@{}} 
 \toprule
 & \multicolumn{2}{c}{{Evaluation Algorithm}}\\
 Model & IWAE & ExVAE & ExVAE (250k)\\
 \midrule
 {VAE + Gaussian Prior} & {$-84.49$} & {$-83.62$} & {$-83.34\pm 0.002$} \\ 
 {VAE + VampPrior} & {$-82.43$} & {$-82.45$} & {$-82.27\pm 0.004$}\\
 {Exemplar VAE} & {$-82.03$} & {$-82.43$} & {$-82.25\pm 0.009$}\\[1ex] 

\bottomrule
\end{tabular}
\caption{Log likelihood lower bounds on the test set (in nats) evaluated based on Exemplar VAE and classical IWAE both with 5000 samples.}
\label{table:ablation-efficent-version}
\end{table}
}

\vspace*{-.1cm}
\subsection{Generative Data Augmentation}
\vspace*{-.1cm}

Finally, we ask whether Exemplar VAE is effective in generating augmented data to improve supervised learning.
Recent generative models have achieved impressive sample quality and diversity, but limited success in improving 
discriminative models. 
Class-conditional models were used to generate training data, but with marginal gains \cite{ravuri2019classification}. 
Techniques for optimizing geometric augmentation policies
\cite{cubuk2019autoaugment,lim2019fast,hataya2019faster} 
and adversarial perturbations \cite{goodfellow2014explaining, miyato2018virtual} were 
more successful for  classification.

Here we use the original training data as exemplars, generating extra samples from Exemplar VAE.
Class labels of source exemplars are transferred to corresponding generated images, and a combination of
real and generated data is used for supervised learning.  Each training iteration involves 3 steps:
\begin{enumerate}[topsep=0pt, partopsep=0pt, leftmargin=13pt, parsep=0pt, itemsep=3pt]
    \item Draw a minibatch ${X} \!=\! \{(\x_i,\ y_i)\}_{i=1}^{B}$ from training data.
    \item For each $\x_i \in {X}$, draw $\z_i \sim r_\phi(\z \mid \x_i)$, 
    and then set $\tilde{\x}_i = \bmu_\phi(\x \mid \z_i)$,
    which inherits the class label $y_i$.
    This yields a synthetic minibatch $\tilde{X} = \{(\tilde{\x}_i,\  y_i)\}_{i=1}^{B}$.
    \item Optimize the weighted cross entropy:
$ \mathcal{\ell} = -\sum_{i=1}^{B}
\Big[
\lambda \log p_{\theta}(y_i \!\mid\! \x_i) + (1\!-\!\lambda)\log p_{\theta}(y_i \!\mid\! \tilde{\x}_i)
\Big]
$
\end{enumerate}

For VAE with Gaussian prior and VampPrior we sampled from variational posterior instead of $r_\phi$. We train MLPs with ReLU activations and two hidden layers of 1024 or 8192 units on MNIST and Fashion MNIST. 
We leverage label smoothing~\cite{szegedy2016rethinking} with a  parameter of $0.1$.
The Exemplar VAEs used for data augmentation have fully connected layers and are not trained with class labels.


Fig.\  \ref{fig:lambda} shows Exemplar VAE is more effective than other VAEs for data augmentation.
Even small amounts of generative data augmentation improves classifier accuracy.
A classifier trained solely on synthetic data achieves better error rates than one trained on the original data. 
Given $\lambda=0.4$ on MNIST and $\lambda=0.8$ on Fashion MNIST,
we train 10 networks on the union of training and validation sets and report average test errors.
On permutation invariant MNIST, Exemplar VAE augmentations achieve an
average error rate of $0.69\%$.
Tables \ref{table:mnist_augmentation} and \ref{table:fashion_augmentation} summarize 
the results in comparison with previous work. Ladder Networks~\cite{sonderby2016ladder} 
and Virtual Adversarial Training \cite{miyato2018virtual} report error rates of $0.57\%$ and $0.64\%$ on MNIST,
using deeper architectures and more complex training procedures.

\begin{table*}[h]
\vspace*{-0.1cm}
\begin{minipage}[h!][][t]{0.46\textwidth}
\centering
\small
 \begin{tabular}{@{\hspace*{.1cm}}l@{}c@{\hspace*{.25cm}}l@{\hspace*{.05cm}}} 
 \toprule
 Method & Hidden layers & Test error\\
 \midrule
 {Dropout} \cite{srivastava2014dropout} & $3\!\times\!1024$ &{$1.25$} \\ 
 {Label smoothing} \cite{pereyra2017regularizing} & $2\!\times\!1024$ & {$1.23 \scriptsize{\,\pm 0.06}$}\\
 {Dropconnect} \cite{wan2013regularization} & $2\!\times\!800$ &{$1.20$} \\
 {VIB} \cite{alemi2016deep} & $2\!\times\!1024$ & {$1.13$}\\
 {Dropout + MaxNorm} \cite{srivastava2014dropout} & $2\!\times\!8192$ & {$0.95$}\\
 {MTC} \cite{rifai2011manifold} &
 $2\!\times\!2000$ & {$0.81$}\\
 {DBM + DO fine.} \cite{srivastava2014dropout} &
 $500{\text -}500{\text -}2K$ & {$0.79$}\\
 \midrule
 {Label Smoothing (LS)}           & $2\!\times\!1024$ & {$1.23
 \scriptsize{\,\pm 0.01}$}\\ 
 {LS+Exemplar VAE Aug.} & $2\!\times\!1024$ & {$0.77 \scriptsize{\,\pm 0.01}$}\\
 {Label Smoothing}           & $2\!\times\!8196$ & {$1.17
 \scriptsize{\,\pm 0.01}$}\\ 
 {LS+Exemplar VAE Aug.} & $2\!\times\!8192$ & {$\mathbf{0.69} \scriptsize{\,\pm 0.01}$}\\
\bottomrule\\[-.5cm]
\end{tabular}
\caption{\small
Test error (\%) on permutation invariant MNIST from 
\cite{srivastava2014dropout,pereyra2017regularizing,wan2013regularization,alemi2016deep,rifai2011manifold},
and our results with and without generative data augmentation. 
}
\label{table:mnist_augmentation}
\end{minipage}
\hspace{.4cm}
\begin{minipage}[h!][][t]{0.52\textwidth}
\vspace*{.25cm}
\centering
\small
 \begin{tabular}{@{\hspace*{.1cm}}l@{}c@{}c@{\hspace*{.05cm}}} 
 \toprule
 Method & Hidden layers & Test error\\
 \midrule
 {Label Smoothing}           & $2\!\times\!1024$ & {$8.96
 \scriptsize{\,\pm 0.04}$}\\ 
 {LS+Exemplar VAE Aug.} & $2\!\times\!1024$ & {$8.46 \scriptsize{\,\pm 0.04}$}\\
 {Label Smoothing}           & $2\!\times\!8196$ & {$8.56
 \scriptsize{\,\pm 0.03}$}\\ 
 {LS+Exemplar VAE Aug.} & $2\!\times\!8192$ & {$\mathbf{8.16} \scriptsize{\,\pm 0.03}$}\\
\bottomrule
\end{tabular}
\caption{\small Test error (\%) on permutaion invariant Fashion MNIST. 
}
\label{table:fashion_augmentation}
%
\vspace*{0.2cm}
\includegraphics[width=0.85\linewidth]{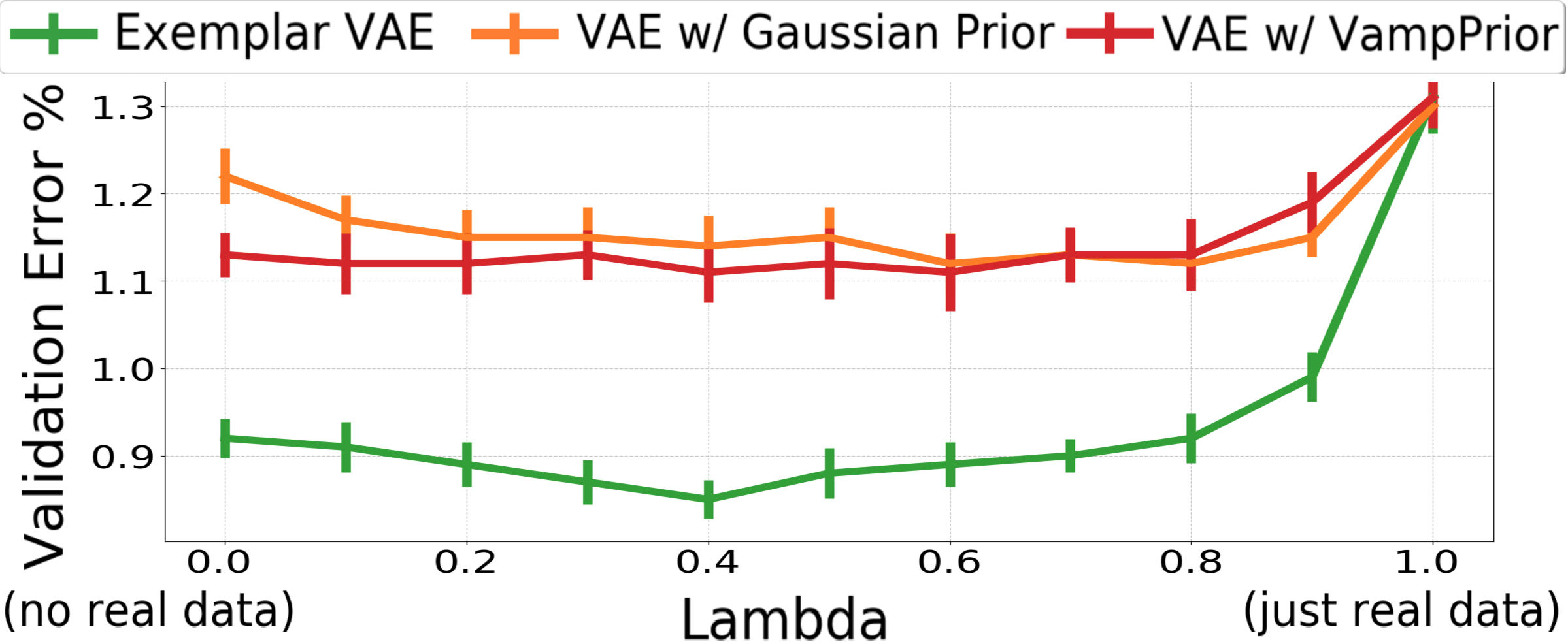}
\captionof{figure}{
\small MNIST validation error versus $\lambda$, which controls 
the relative balance of real and augmented data,  
for  different generative models.
}
\label{fig:lambda}
\vspace*{-0.25cm}

\end{minipage}
\vspace*{-0.1cm}
\end{table*}



\vspace*{-.1cm}
\section{Conclusion}
\vspace*{-.1cm}

We develop a framework for exemplar based generative modeling  called
the Exemplar VAE. 
We present two effective
regularization techniques for Exemplar VAEs, and an efficient learning algorithm 
based on approximate nearest neighbor search.  The effectiveness of the Exemplar VAE on
density estimation, representation learning, and data augmentation for
supervised learning is demonstrated.
The development of Exemplar VAEs opens up interesting future research directions such as
application to NLP (cf.\ \cite{guu2018generating}) and other discrete 
data, further exploration of unsupervised data augmentation, and extentions to other generative models such as Normalizing
Flows and GANs.



\section*{Broader Impact Statement}
\vspace*{-0.15cm}

The ideas described in our paper concern the development of a new fundamental class of unsupervised
learning algorithm, rather than an application per se. One important property of the method stems
from it’s non-parametric form, i.e., as an exemplar-based model. As such, rather than having the
"model" represented solely in the weights of an amorphous non-linear neural network, in our case
much of the model is expressed directly in terms of the dataset of exemplars. As such, the model is
somewhat more interpretable and may facilitate the examination or discovery of bias, which has
natural social and ethical implications. Beyond that, the primary social and ethical implications will
derive from the way in which the algorithm is applied in different domains.

\section*{Funding Disclosure}
\vspace*{-0.15cm}

This research was supported in part by an NSERC Discovery Grant to DJF, and by Province of Ontario, the Government of Canada, through NSERC and CIFAR, and companies sponsoring the Vector Institute.
\vspace*{-.1cm}
\section*{Acknowledgement}
\vspace*{-.1cm}

We are extremely grateful to Micha Livne, Will Grathwohl, and Kevin Swersky for extensive discussions.
We thank Alireza Makhzani, Kevin Murphy, Abhishek Gupta, and Alex Alemi for useful discussions and
Diederik Kingma, Chen Li, Danijar Hafner, and David Duvenaud for their 
valuable feedback on an initial draft of this paper.

\newpage



\bibliographystyle{plain}
\bibliography{paper}

\appendix
\newpage


\section{Exemplar VAE samples}

\begin{figure}[H]
\small
\begin{center}
\begin{tabular}{ccc}
    \includegraphics[width=.32\linewidth]{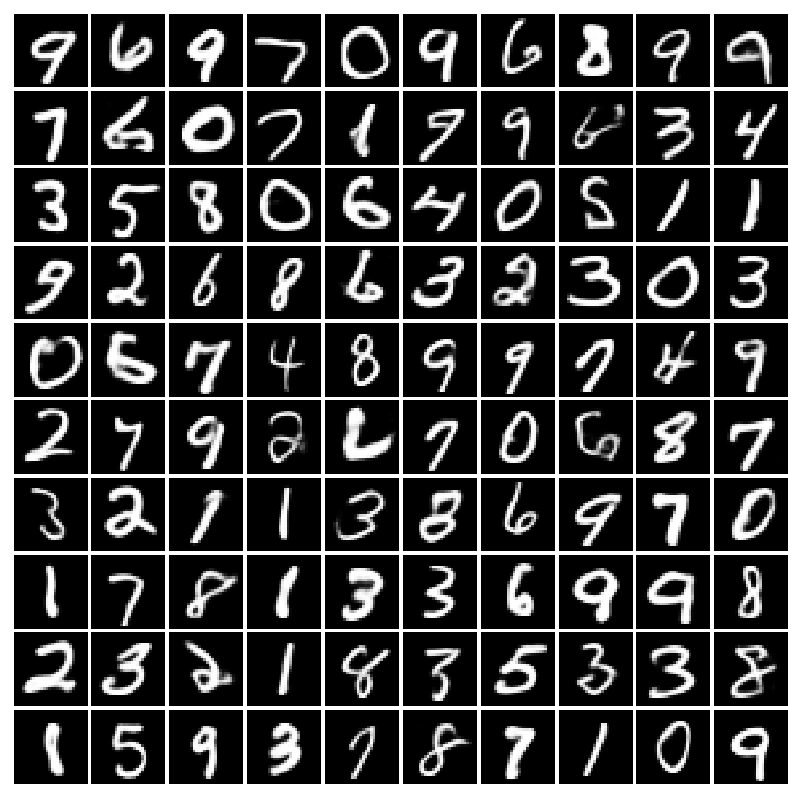} &
    \includegraphics[width=.32\linewidth]{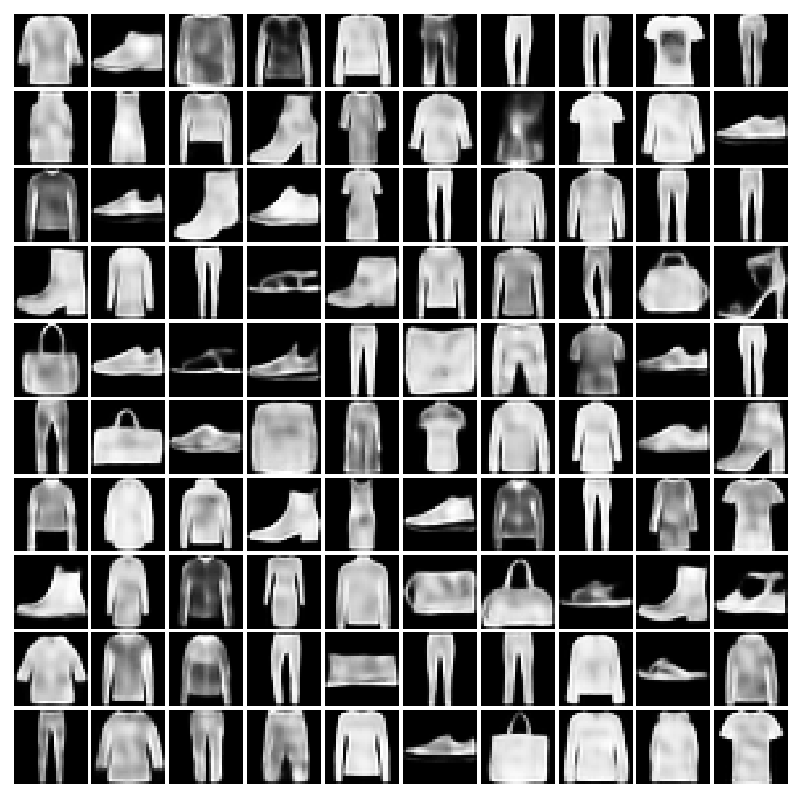} &
    \includegraphics[width=.32\linewidth]{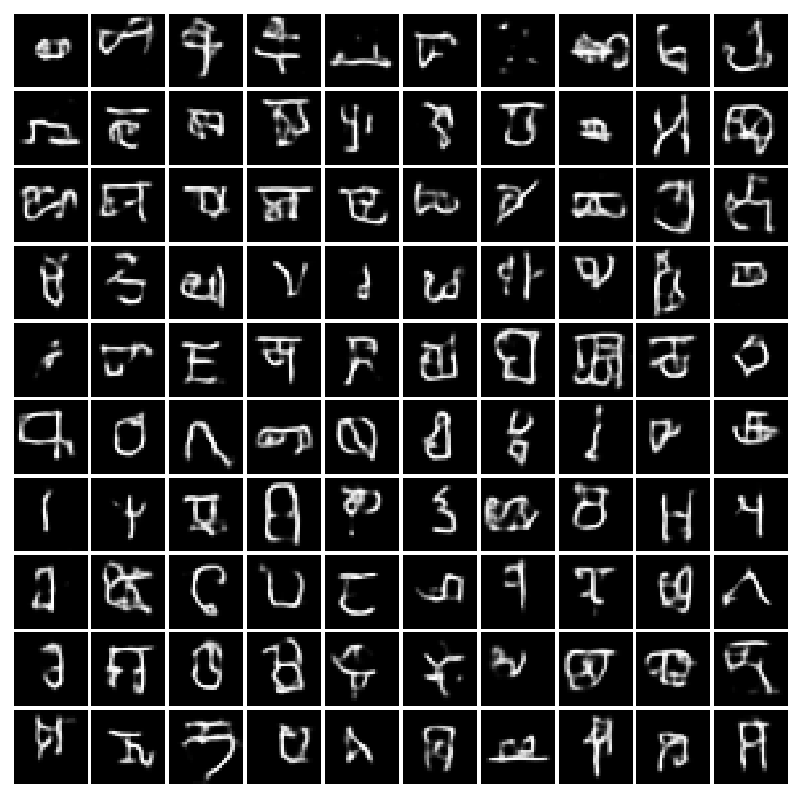}\\
    {MNIST} & {Fashion MNIST} & {Omniglot} \\[.3cm]
    \multicolumn{3}{c}{
     \includegraphics[width=1\linewidth]{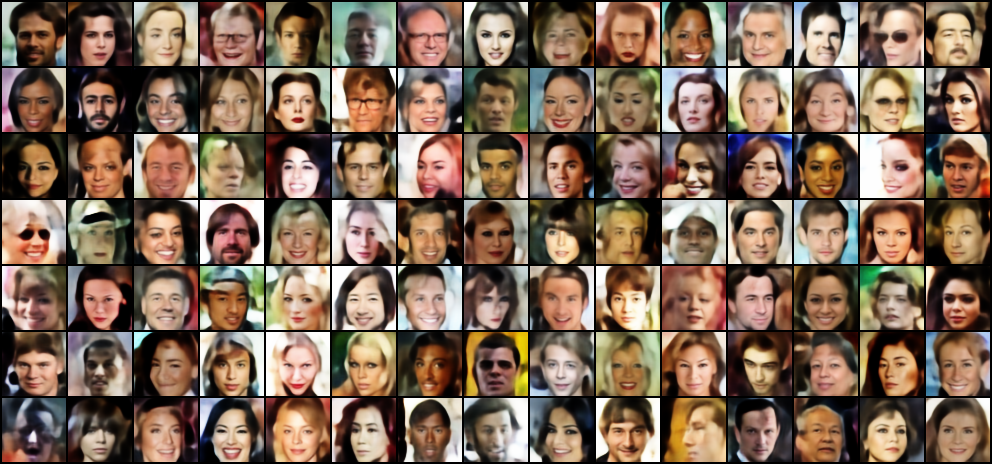}}\\
    \multicolumn{3}{c}{CelebA}
    \end{tabular}
    \end{center}
    \vspace*{-0.3cm}
  \caption{{Random samples drawn from Exemplar VAEs trained on different datasets.}}
\end{figure}

\vfill

\pagebreak

\section{Exemplar conditioned samples}

\begin{figure*}[!h]
\small
\begin{center}
\begin{tabular}{@{}
c@{\hspace{0.01\linewidth}}c@{\hspace{0.01\linewidth}}
c@{\hspace{0.01\linewidth}}c@{\hspace{0.01\linewidth}}
c@{\hspace{0.01\linewidth}}c@{}
}
\includegraphics[width=0.16\linewidth]{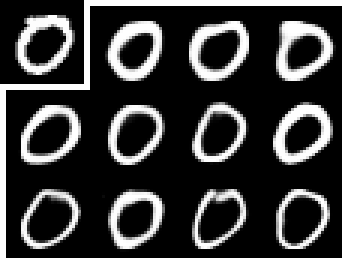} &
\includegraphics[width=0.16\linewidth]{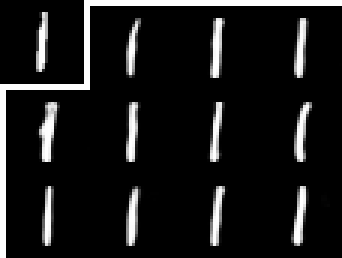} &
\includegraphics[width=0.16\linewidth]{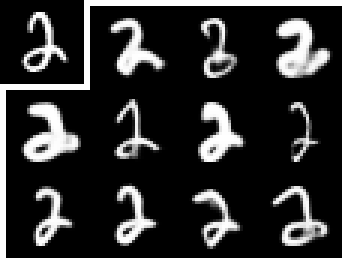} &
\includegraphics[width=0.16\linewidth]{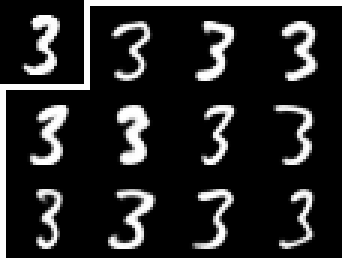}&
\includegraphics[width=0.16\linewidth]{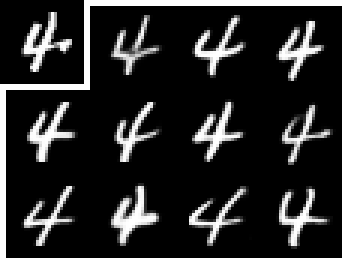} &
\includegraphics[width=0.16\linewidth]{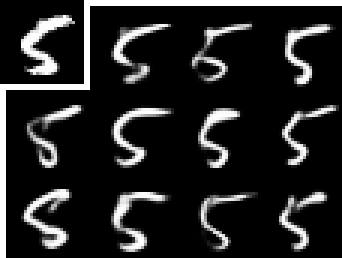} \\
\includegraphics[width=0.16\linewidth]{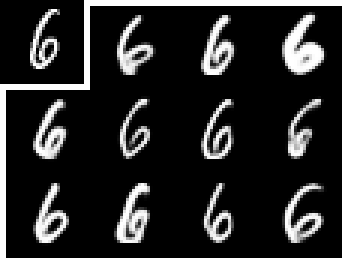} &
\includegraphics[width=0.16\linewidth]{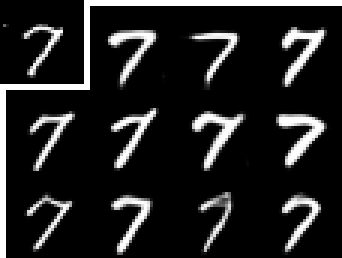} &
\includegraphics[width=0.16\linewidth]{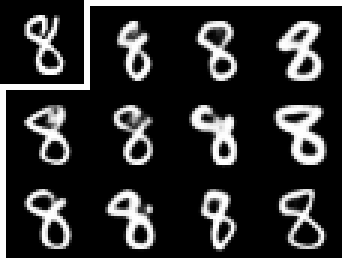} &
\includegraphics[width=0.16\linewidth]{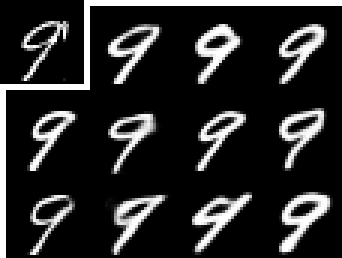} &
\includegraphics[width=0.16\linewidth]{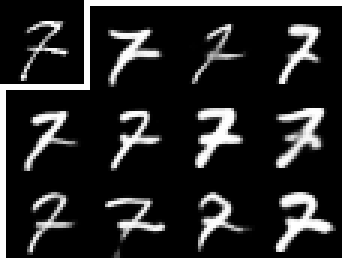} &
\includegraphics[width=0.16\linewidth]{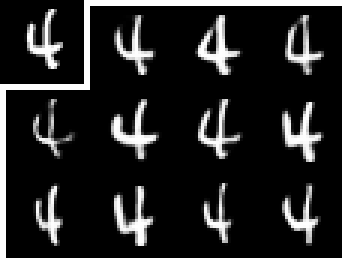}\\ \\
\includegraphics[width=0.16\linewidth]{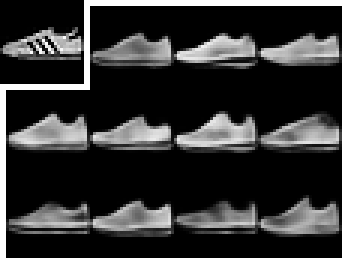} &
\includegraphics[width=0.16\linewidth]{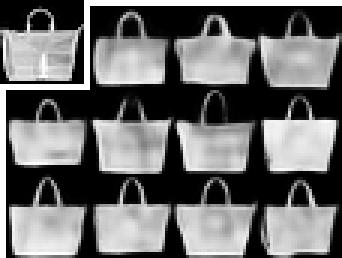} &
\includegraphics[width=0.16\linewidth]{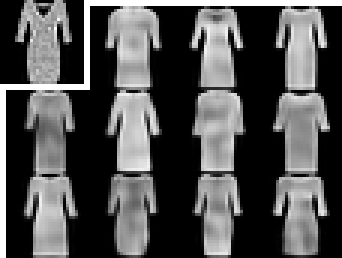} &
\includegraphics[width=0.16\linewidth]{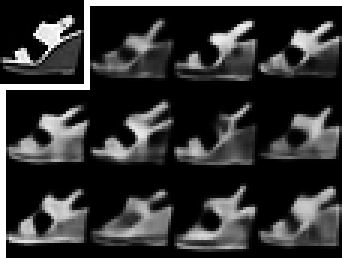}&
\includegraphics[width=0.16\linewidth]{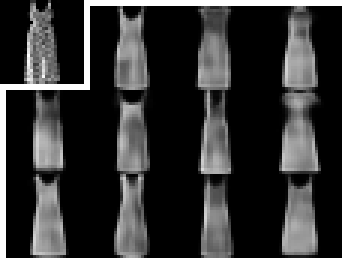} &
\includegraphics[width=0.16\linewidth]{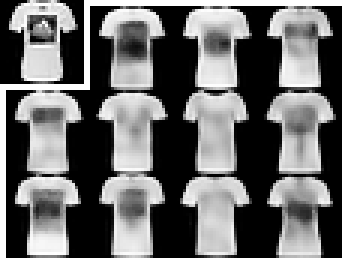} \\
\includegraphics[width=0.16\linewidth]{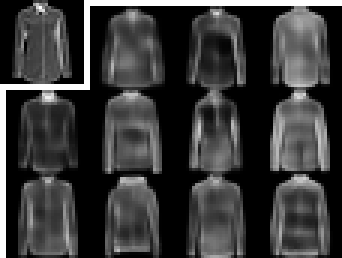} &
\includegraphics[width=0.16\linewidth]{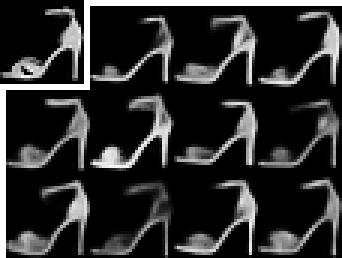} &
\includegraphics[width=0.16\linewidth]{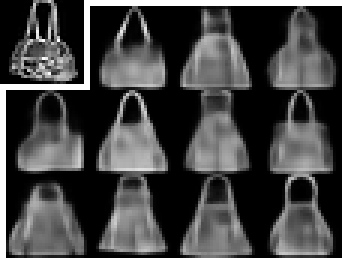} &
\includegraphics[width=0.16\linewidth]{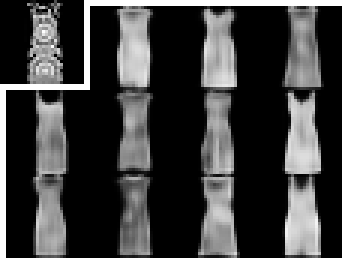} &
\includegraphics[width=0.16\linewidth]{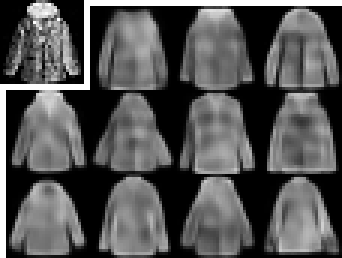} &
\includegraphics[width=0.16\linewidth]{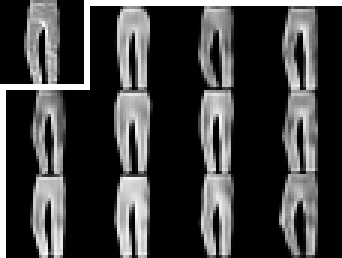}\\\\
\includegraphics[width=0.16\linewidth]{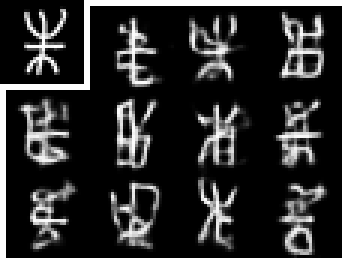} &
\includegraphics[width=0.16\linewidth]{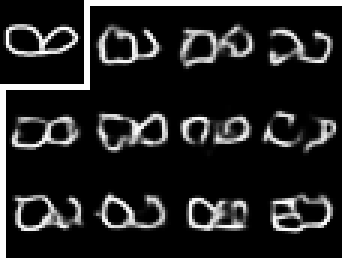} &
\includegraphics[width=0.16\linewidth]{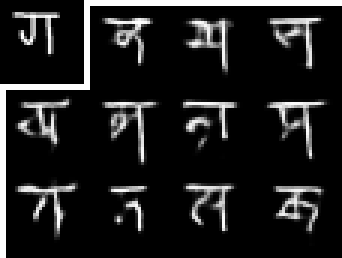} &
\includegraphics[width=0.16\linewidth]{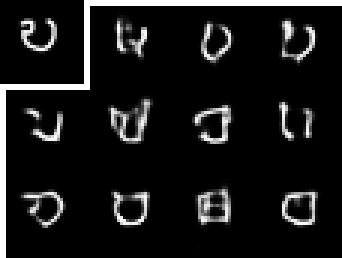}&
\includegraphics[width=0.16\linewidth]{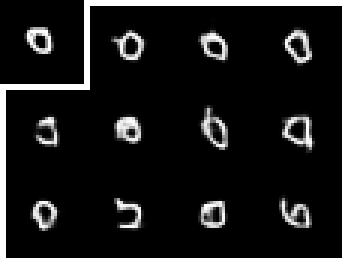} &
\includegraphics[width=0.16\linewidth]{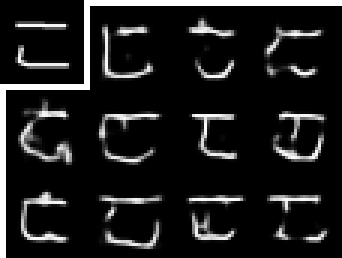} \\
\includegraphics[width=0.16\linewidth]{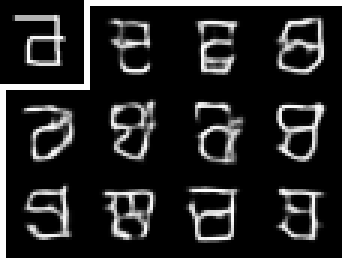} &
\includegraphics[width=0.16\linewidth]{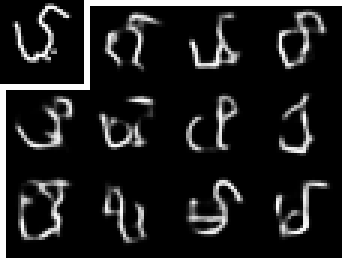} &
\includegraphics[width=0.16\linewidth]{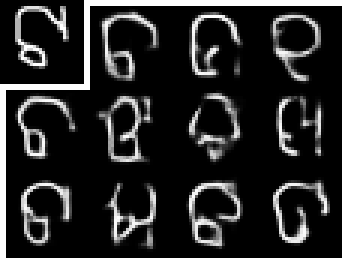} &
\includegraphics[width=0.16\linewidth]{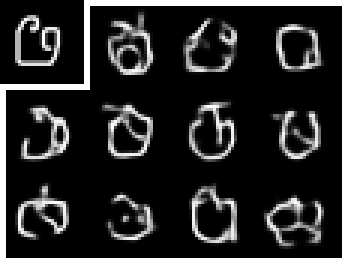} &
\includegraphics[width=0.16\linewidth]{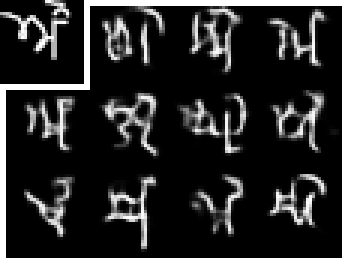} &
\includegraphics[width=0.16\linewidth]{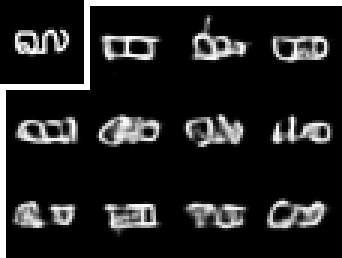}\\\\
\end{tabular}
\begin{tabular}{@{}
c@{\hspace{0.01\linewidth}}c@{\hspace{0.01\linewidth}}c@{\hspace{0.01\linewidth}}c@{}
}
\includegraphics[width=0.24\linewidth]{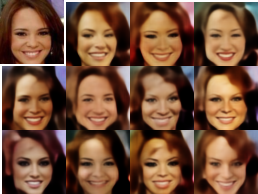} &
\includegraphics[width=0.24\linewidth]{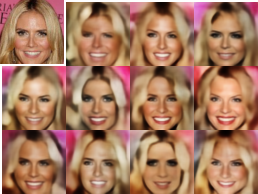} &
\includegraphics[width=0.24\linewidth]{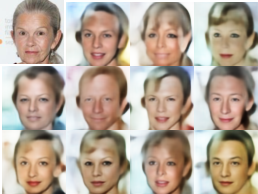} &
\includegraphics[width=0.24\linewidth]{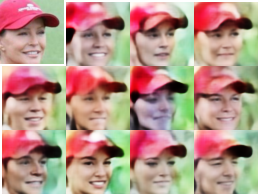} \\
\includegraphics[width=0.24\linewidth]{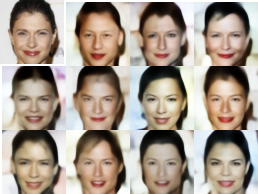} &
\includegraphics[width=0.24\linewidth]{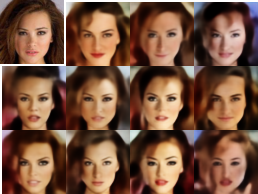} &
\includegraphics[width=0.24\linewidth]{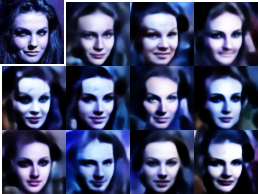} &
\includegraphics[width=0.24\linewidth]{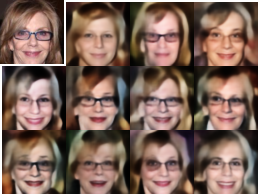} \\
\includegraphics[width=0.24\linewidth]{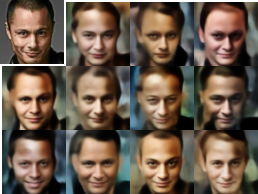} &
\includegraphics[width=0.24\linewidth]{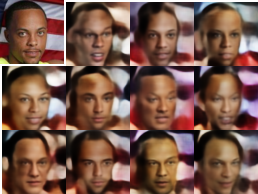} &
\includegraphics[width=0.24\linewidth]{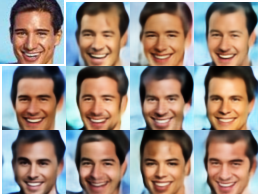} &
\includegraphics[width=0.24\linewidth]{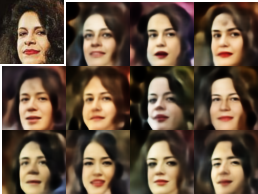}\\
\end{tabular}
\end{center}
\caption{\small{Given the input exemplar on the top left of each plate, $11$ exemplar conditioned
samples using Exemplar VAE are generated and shown.}
\label{figure:mnist-generation}
}
\end{figure*}

\pagebreak

\comment{

\begin{figure*}[H]
\small
\begin{center}
\begin{tabular}{@{}
c@{\hspace{0.01\linewidth}}c@{\hspace{0.01\linewidth}}
c@{\hspace{0.01\linewidth}}c@{\hspace{0.01\linewidth}}
}
\includegraphics[width=0.15\linewidth]{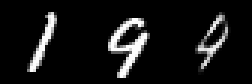} &
\includegraphics[width=0.15\linewidth]{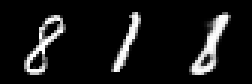} &
\includegraphics[width=0.15\linewidth]{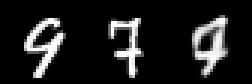} &
\includegraphics[width=0.15\linewidth]{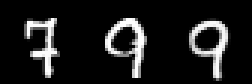} \\
\includegraphics[width=0.15\linewidth]{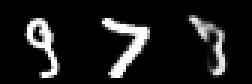} &
\includegraphics[width=0.15\linewidth]{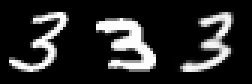} &
\includegraphics[width=0.15\linewidth]{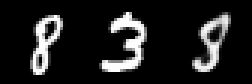} &
\includegraphics[width=0.15\linewidth]{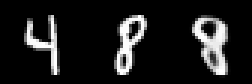} \\
\includegraphics[width=0.15\linewidth]{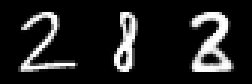} &
\includegraphics[width=0.15\linewidth]{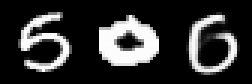} &
\includegraphics[width=0.15\linewidth]{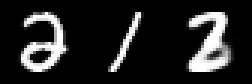} &
\includegraphics[width=0.15\linewidth]{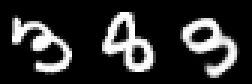} 
\end{tabular}
\end{center}
\vspace*{-0.25cm}
\caption{\small{First half of latent space of the first image combined with the second half of the second image creates the thrid image.}
\label{figure:mnist-generation}
\vspace*{-.3cm}
}
\end{figure*}

\begin{figure*}[H]
\small
\begin{center}
\begin{tabular}{@{}
c@{\hspace{0.01\linewidth}}c@{\hspace{0.01\linewidth}}
c@{\hspace{0.01\linewidth}}c@{\hspace{0.01\linewidth}}
}
\includegraphics[width=0.2\linewidth]{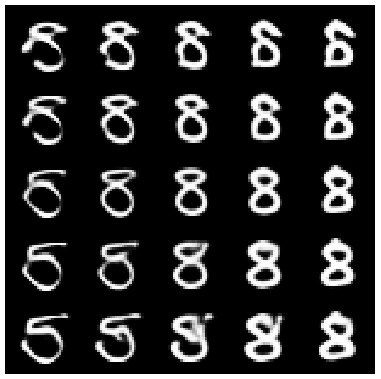} &
\includegraphics[width=0.2\linewidth]{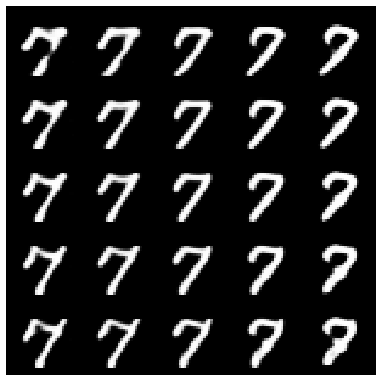} &
\includegraphics[width=0.2\linewidth]{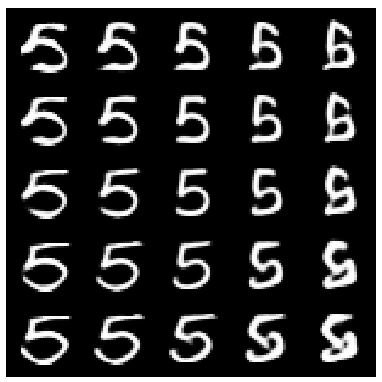}\\
\includegraphics[width=0.2\linewidth]{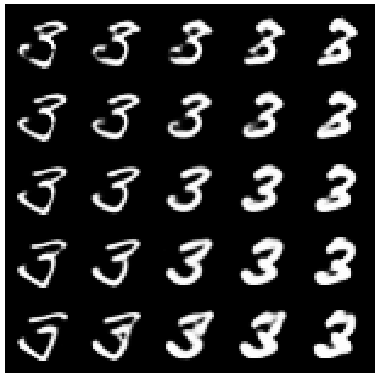} &
\includegraphics[width=0.2\linewidth]{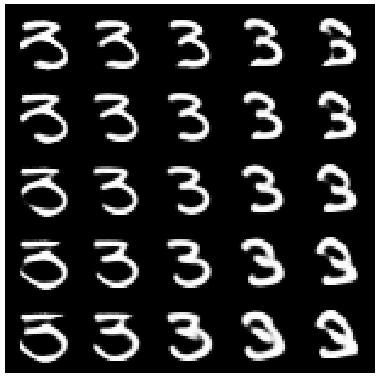} &
\includegraphics[width=0.2\linewidth]{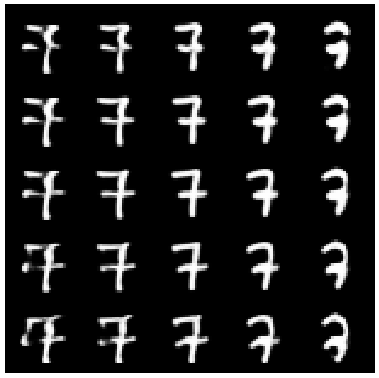}\\
\includegraphics[width=0.2\linewidth]{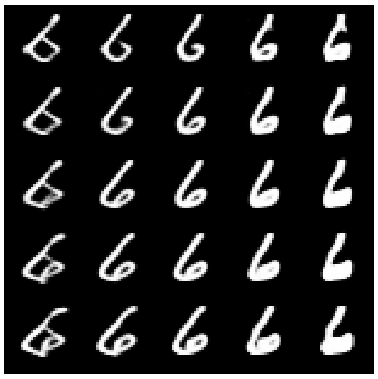} &
\includegraphics[width=0.2\linewidth]{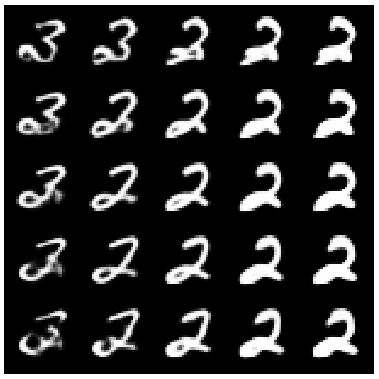} &
\includegraphics[width=0.2\linewidth]{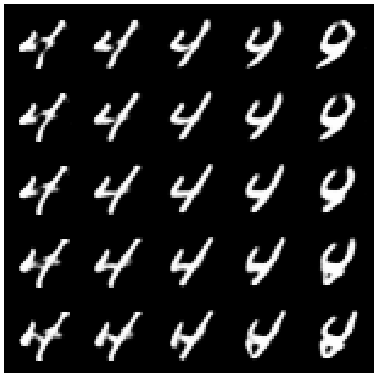}
\end{tabular}
\end{center}
\vspace*{-0.25cm}
\caption{\small{Grid interpolation in the latent space given the image in center of each 5 by 5 image.}
\label{figure:mnist-generation}
\vspace*{-.3cm}
}
\end{figure*}
}

\vspace{-0.5cm}
\section{Retrieval Augmented Training}
\vspace{-0.5cm}
\begin{algorithm}[H]
\caption{\label{alg:main}
}
\begin{algorithmic}
    \STATE \textbf{Input:} Training dataset $\mathcal{X}=\{ \x_n\}_{n=1}^N$\\
    \STATE \textbf{Define} Cache:
    \STATE \qquad initialize cache = []
    \STATE \qquad $\mathrm{insert}(i, \bm c)$: insert value $\bm c$ with index $i$ into cache 
    \STATE \qquad $\mathrm{update}(i, \bm c)$: update the value of index $i$ to $\bm c$
    \STATE \qquad $\mathrm{kNN}(\bm c)$: return indices of kNNs of $\bm c$ in cache\\
    \STATE \textbf{for} $n$ ~\textbf{in}~ $\{1, \ldots, N\}$ \textbf{do} $\mathrm{Cache.insert}(n, \bmu_{\phi}(\x_n))$\\
    \STATE \textbf{for} epoch ~\textbf{in}~ $\{1, \ldots, L\}$ \textbf{do}\\
    \STATE \quad\textbf{for} $i$ ~\textbf{in}~ $\{1, \ldots, N\}$ \textbf{do}\\
        \STATE \qquad $\pi \sim \Pi^{N,i}_{M}$ to obtain a set of $M$ exemplar indices\\
        \STATE \qquad $\bmu_{i}, \Lambda_{i} = \bmu_{\phi}(\x_i), \Lambda_{\phi}(\x_i) $\\

        \STATE \qquad $\bm \epsilon \sim \mathcal{N}(0, I_{d_z\times d_z})$\\
        \STATE \qquad $\z = \bmu_{i} +\Lambda_i^{1/2}\bm \epsilon$
        \STATE \qquad $\mathrm{kNN} = \mathrm{Cache.kNN}(\bmu_i) \cap \ \pi$\\
        \STATE \qquad \textbf{for} $j$ ~\textbf{in}~ $\mathrm{kNN}$ \textbf{do} $\bmu_{j} = \bmu_{\phi}(\x_{j})$\\
        \STATE \qquad $m(\z) = \frac{1}{M}\sum_{j\in \mathrm{kNN}} \mathcal{N}(\z \!\mid\! \bmu_{j}, \sigma^2)$\\
        \STATE \qquad $\mathrm{ELBO} \!=\! \log p_{\theta}(\x \!\mid\! \z) \!-\! \log \mathcal{N}(\z \!\mid\! \bmu_i, \Lambda_i)
\!+\! \log r(\z)$
        \STATE \qquad Gradient ascend on $\mathrm{ELBO}$ to update $\phi$, $\theta$, and $\sigma^2$\\
        \STATE \qquad $\mathrm{Cache.update}(i, \bmu_i)$
        \STATE \qquad \textbf{for} $j$ ~\textbf{in}~ $\mathrm{kNN}$ \textbf{do} $\mathrm{Cache.update}(j$, $\bmu_j)$
\end{algorithmic}
\end{algorithm}
\vspace*{-0.2cm}

\section{Number of Active Dimensions in the Latent Space}

The problem of posterior collapse~\cite{bowman2015generating,lucas2019don}, resulting in a number of inactive dimensions in the latent space of a VAE. We investigate this phenomena by counting the number of active dimensions based on a metric proposed by Burda et. al~\cite{burda2015importance}. This metric computes the variance of the mean of the latent encoding of the data points in each dimension of the latent space, $\mathrm{Var}(\mu_{\phi}(\x)_i)$, where $\x$ is sampled from the dataset. If the computed variance is above a certain threshold, then that dimension is considered active. The proposed threshold by \cite{bauer2018resampled} is $0.01$ and we use the same value. We observe that the Exemplar VAE has the largest number of active dimensions in all cases except one. In the case of ConvHVAE and PixelSNAIL, the gap between Exemplar VAE and other methods is more considerable.
\begin{table*}[!h]
    \centering
    \begin{tabular}{r c c c}
    \toprule
    & \multicolumn{3}{c}{Number of active dimensions out of $40$}\\
    Model & {Dynamic MNIST} & {Fashion MNIST} & {Omniglot}\\
    \midrule
    VAE w/ Gaussian prior &$24.0${\footnotesize $\pm0.63$}&$26.0${\footnotesize $\pm1.1$}&$35.2${\footnotesize $\pm0.4$}\\
    VAE w/ Vampprior  &$27.6${\footnotesize $\pm1.36$}&$35.25${\footnotesize $\pm1.3$}&{\boldmath$40.0$}{\footnotesize $\pm0.0$} \\
    Exemplar VAE &{\boldmath$29.4$}{\footnotesize $\pm0.49$}&{\boldmath $36.0$}{\footnotesize $\pm1.41$}&{\boldmath$40.0$}{\footnotesize $\pm0.0$}\\
    \midrule
    HVAE w/ Gaussian prior &$15.0${\footnotesize $\pm0.63$}&$12.4${\footnotesize $\pm0.8$}&$24.8${\footnotesize $\pm1.83$}\\
    HVAE w/ VampPrior &$20.4${\footnotesize $\pm0.49$}&$23.2${\footnotesize $\pm1.47$}&{\boldmath $39.0$}{\footnotesize $\pm0.89$}\\
    Exemplar HVAE &{\boldmath $21.6$}{\footnotesize $\pm0.49$}&{\boldmath$28.6$}{\footnotesize $\pm0.8$}&$38.6${\footnotesize $\pm1.5$}\\
    \midrule
    ConvHVAE w/ Gaussian prior &$19.8${\footnotesize $\pm2.93$}&$15.4${\footnotesize $\pm2.65$}&$39.2${\footnotesize $\pm1.6$}\\
    ConvHVAE w/ VampPrior &$19.0${\footnotesize $\pm1.55$}&$19.25${\footnotesize $\pm0.83$}&$39.8${\footnotesize $\pm0.4$}\\
    Exemplar ConvHVAE &{\boldmath $25.8$}{\footnotesize $\pm3.66$}&{\boldmath $33.6$}{\footnotesize $\pm7.86$}&{\boldmath$40.0$}{\footnotesize $\pm0.0$}\\
    \midrule
    PixelSNAIL w/ Gaussian prior &$4.6${\footnotesize $\pm0.36$}&$2.4${\footnotesize $\pm0.22$}&$0.0${\footnotesize $\pm0.0$}\\
    PixelSNAIL w/ VampPrior &$17.2${\footnotesize $\pm1.39$}&$30.6${\footnotesize $\pm0.73$}&$9.50${\footnotesize $\pm4.48$}\\
    Exemplar PixelSNAIL &{\boldmath $25.8$}{\footnotesize $\pm0.66$}&{\boldmath $37.2$}{\footnotesize $\pm0.87$}&{\boldmath$25.5$}{\footnotesize $\pm1.82$}\\
    \bottomrule
    \end{tabular}
    \caption{The number of active dimensions computed based on a metric proposed by Burda et. al~\cite{burda2015importance}. This metric considers a latent dimension active if the variance of its mean over the dataset is higher than $0.01$. For hierarchical architectures the reported number is for the $\z_2$ which is the highest stochastic layer.}
    \label{tab:posterior_collapse}
\vspace*{-.5cm}
\end{table*}

\newpage
\section{CelebA Quantitative Results}


\begin{table*}[!h]
    \centering
    \begin{tabular}{c c}
    \toprule
         Model & bits per dim \\
         \midrule
         VAE w/ Gaussian Prior& $5.825$\\
         Exemplar VAE& {\boldmath$5.780$ } \\
         \bottomrule
    \end{tabular}
    \caption{Numerical Evaluations for CelebA}
    \label{tab:my_label}
\end{table*}

\section{Derivation of Eqn.\ (5)}
\vspace{-0.5cm}
\begin{align}
    \log p(\x; X, \theta, \phi)  &~=~
    \log \sum_{n=1}^N \frac{1}{N}\int_z {r_\phi(\z \mid \x_n) \,p_\theta(\x \mid \z)}\, d\z\\
     &~=~
    \log \int_z p_\theta(\x \mid \z) \sum\nolimits_{n=1}^N \frac{1}{N} {r_\phi(\z \mid \x_n) \,}\, d\z\\
     &~=~
    \log \int_z \frac{q_\phi(\z \mid \x)  p_\theta(\x \mid \z) \sum\nolimits_{n=1}^N \frac{1}{N} {r_\phi(\z \mid \x_n) \,}}{q_\phi(\z|\x)}\, d\z\\
    &~\ge~ \underbrace{\mathop{\expected}_{q_{\phi}(\z \mid \x)} \!\!\!
    \log  p_{\theta}(\x\!\mid\!\z)}_{\mathrm{reconstruction}} - \!\!\underbrace{\mathop{\expected}_{q_{\phi}(\z \mid \x)}
    \log \frac{q_\phi(\z \mid \x)}{\sum\nolimits_{n=1}^N r_\phi(\z \mid \x_n)/N}}_{\mathrm{KL~term}} \\
    &~=~ O(\theta, \phi; \x, X).
\end{align}


\vspace{-0.7cm}
\section{Iterative generation}
\begin{figure}[!h]
\begin{subfigure}{\linewidth}
    \includegraphics[width=\linewidth]{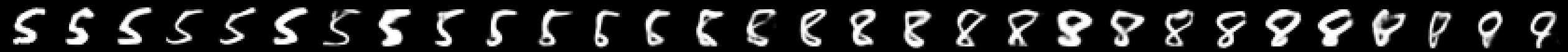}
\end{subfigure}
\begin{subfigure}{\linewidth}
    \includegraphics[width=\linewidth]{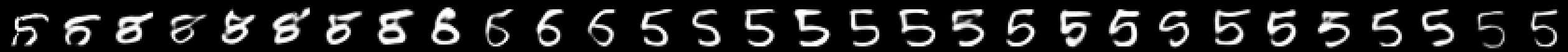}
\end{subfigure}
\begin{subfigure}{\linewidth}
    \includegraphics[width=\linewidth]{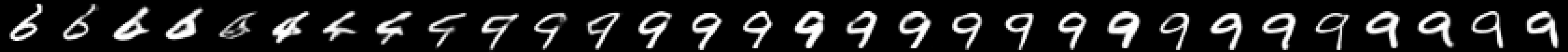}
\end{subfigure}
\begin{subfigure}{\linewidth}
    \includegraphics[width=\linewidth]{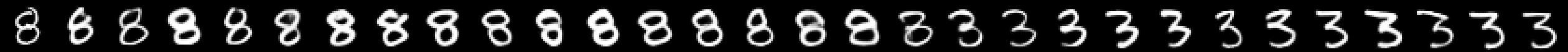}
\end{subfigure}
\begin{subfigure}{\linewidth}
    \includegraphics[width=\linewidth]{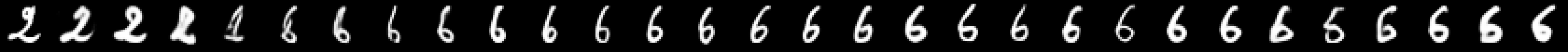}
\end{subfigure}
\begin{subfigure}{\linewidth}
    \includegraphics[width=\linewidth]{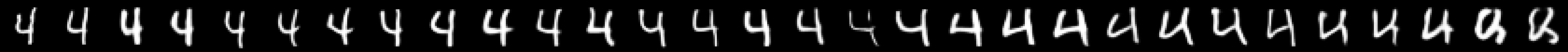}
\end{subfigure}
\caption{Iterative generation starting from a training data point. Samples generated from an Exemplar VAE starting from a training data point, and then reusing the generated data as exemplars for the next round of generation (left to right). }
\label{fig:cyclic_generation}
\end{figure}

\begin{figure}[!h]
\begin{subfigure}{\linewidth}
    \includegraphics[width=\linewidth]{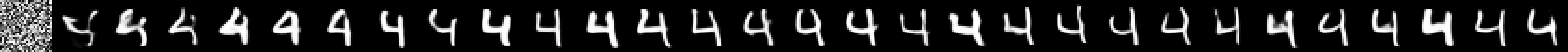}
\end{subfigure}
\begin{subfigure}{\linewidth}
    \includegraphics[width=\linewidth]{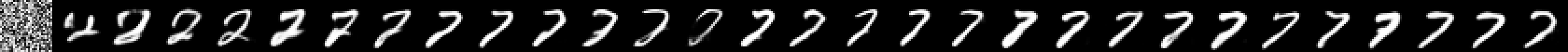}
\end{subfigure}
\begin{subfigure}{\linewidth}
    \includegraphics[width=\linewidth]{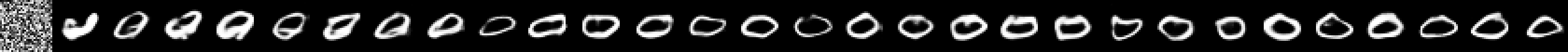}
\end{subfigure}
\begin{subfigure}{\linewidth}
    \includegraphics[width=\linewidth]{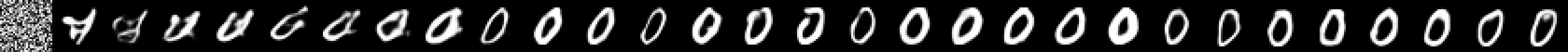}
\end{subfigure}
\begin{subfigure}{\linewidth}
    \includegraphics[width=\linewidth]{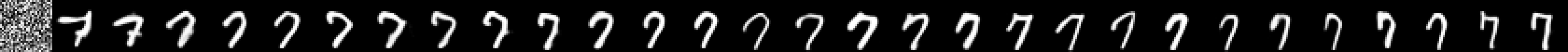}
\end{subfigure}
\begin{subfigure}{\linewidth}
    \includegraphics[width=\linewidth]{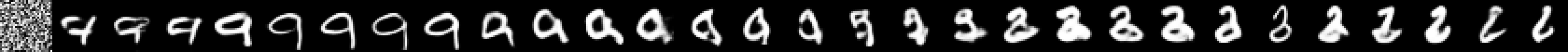}
\end{subfigure}
\caption{Iterative generation starting from a noise input (left to right). }
\label{fig:cyclic_generation_noise}
\end{figure}

The exemplar VAE generates a new sample by stochastically transforming an exemplar. The newly generated data point 
can also be used as an exemplar, and we can repeat this procedure again and again. This kind of generation bears some
similarity to MCMC for sampling from energy-based models. \figref{fig:cyclic_generation} shows how samples evolve and
consistently stay near the manifold of MNIST digits. We can apply the same procedure starting from a noisy input image
as an exemplar. \figref{fig:cyclic_generation_noise} shows that the model is able to quickly transform the noisy images
into samples that resemble real MNIST images.

\newpage
\section{Computation and Memory Complexity}
\vspace{-0.1cm}
The cost of training Exemplar VAE is similar to that of VampPrior, which uses mixture of variational posteriors. When the number of exemplars per minibatch is equal to the number of pseudo-inputs in VampPrior the computational complexity is very similar. 
For example, for ConvHVAE on Omniglot, VampPrior with 1000 pseudo-inputs takes 58s/epoch and Exemplar VAE with a minibatch of 100 and 10 NNs takes 51s/epoch on a single Nvidia T4 GPU (it runs faster because we use an isotropic gaussians in our prior). 
In case of ConvHVAE on MNIST and FashionMNIST VampPrior with 500 pseudo inputs takes 82s/epoch vs 107s/epoch for Exemplar VAE with batch size of 100 and 10 NNs per data point. 
Regarding memory complexity, Exemplar VAE stores low-dimensional latent embeddings.  
By comparison, VampPrior stores pseudo inputs with the same dimentionality as the input data, which can be problematic in case of high dimensional data.

\section{Reconstruction vs. KL}
\vspace{-0.1cm}
\tabref{tab:reconst_kl} shows the value of KL and the reconstruction terms of ELBO,
computed based on a single sample from the variational posterior, averaged across test set.
On non-autoregressive architectures, these numbers show that not
only the exemplar VAE improves the KL term, but also the reconstruction terms are comparable with the VampPrior. On PixelSNAIL, these numbers confirm that Exemplar PixelSNAIL utilize the latent space better. 

\begin{table*}[!h]
    \hspace*{-2cm}
    \centering
    \begin{tabular}{r cc cc cc}
    \toprule
    & \multicolumn{2}{c}{{Dynamic MNIST}} & \multicolumn{2}{c}{{Fashion MNIST}} & \multicolumn{2}{c}{{Omniglot}}\\
    Model & KL & Neg.Reconst. & KL & Neg. Reconst. & KL & Neg. Reconst.  \\ 
    \midrule
    VAE w/ Gaussian prior &$25.54${\footnotesize$\pm0.12$}&$63.06${\footnotesize$\pm0.11$}&$18.38${\footnotesize$\pm0.11$}&$213.21${\footnotesize$\pm0.18$}&$32.97${\footnotesize$\pm0.2$}&$82.3${\footnotesize$\pm0.21$} \\
    VAE w/ VampPrior &$25.14${\footnotesize$\pm0.16$}&{\boldmath$60.79$}{\footnotesize$\pm0.13$}&$18.44${\footnotesize$\pm0.06$}&$211.37${\footnotesize$\pm0.04$}&$34.17${\footnotesize$\pm0.22$}&{\boldmath$79.49$}{\footnotesize$\pm0.18$}  \\
    Exemplar VAE &{\boldmath$24.82$}{\footnotesize$\pm0.22$}&$61.00${\footnotesize$\pm0.13$}&{\boldmath$18.32$}{\footnotesize$\pm0.08$}&{\boldmath $211.10$}{\footnotesize$\pm0.1$}&{\boldmath$32.66$}{\footnotesize$\pm0.27$}&$80.25${\footnotesize$\pm0.62$} \\
    \midrule
    HVAE w/ Gaussian prior &$26.80${\footnotesize$\pm0.13$}&$59.80${\footnotesize$\pm0.11$}&$19.08${\footnotesize$\pm0.05$}&$211.18${\footnotesize$\pm0.14$}&{\boldmath$36.07$}{\footnotesize$\pm0.12$}&$75.96${\footnotesize$\pm0.12$} \\
    HVAE w/ VampPrior &$26.69${\footnotesize$\pm0.1$}&{\boldmath $58.46$}{\footnotesize$\pm0.06$}&$19.27${\footnotesize$\pm0.15$}&{\boldmath $210.04$}{\footnotesize$\pm0.2$}&$38.39${\footnotesize$\pm0.16$}&{\boldmath$72.42$}{\footnotesize$\pm0.34$}\\
    Exemplar HVAE &{\boldmath $26.41$}{\footnotesize$\pm0.17$}&$58.48${\footnotesize$\pm0.16$}&{\boldmath$18.96$}{\footnotesize$\pm0.15$}&$210.40${\footnotesize$\pm0.16$}&$36.76${\footnotesize$\pm0.25$}&$73.35${\footnotesize$\pm0.63$}
    \\
    \midrule
    ConvHVAE w/ Gaussian prior &$26.58${\footnotesize$\pm0.27$}&$57.64${\footnotesize$\pm0.57$}&$20.34${\footnotesize$\pm0.04$}&$208.11${\footnotesize$\pm0.06$}&$38.90${\footnotesize$\pm0.22$}&$67.22${\footnotesize$\pm0.1$}\\
    ConvHVAE w/ VampPrior &$26.57${\footnotesize$\pm0.17$}&$56.18${\footnotesize$\pm0.03$}&$20.65${\footnotesize$\pm0.19$}&{\boldmath $206.64$}{\footnotesize$\pm0.15$}&$38.95${\footnotesize$\pm0.17$}&{\boldmath $66.38$}{\footnotesize$\pm0.3$}\\
    Exemplar ConvHVAE &{\boldmath$26.41$}{\footnotesize$\pm0.25$}&{\boldmath$56.14$}{\footnotesize$\pm0.27$}&{\boldmath$ 20.46$}{\footnotesize$\pm0.23$}&$207.18${\footnotesize$\pm0.38$}&{\boldmath $37.48$}{\footnotesize$\pm0.37$}&$66.62${\footnotesize$\pm0.32$}\\
    \midrule
    PixelSNAIL w/ Gaussian prior &{\boldmath $5.73$}{\footnotesize$\pm0.09$}&$72.92${\footnotesize$\pm0.08$}&{\boldmath $4.68$}{\footnotesize$\pm0.16$}&$219.51${\footnotesize$\pm0.16$}&{\boldmath $0.01$}{\footnotesize$\pm0.0$}&$89.57${\footnotesize$\pm0.07$}\\
    PixelSNAIL w/ VampPrior &$7.17${\footnotesize$\pm0.18$}&$71.23${\footnotesize$\pm0.19$}&$6.07${\footnotesize$\pm0.07$}& $218.00${\footnotesize$\pm0.08$}&$1.07${\footnotesize$\pm0.42$}&$88.54${\footnotesize$\pm0.47$}\\
    Exemplar PixelSNAIL &$10.94${\footnotesize$\pm0.03$}&{\boldmath$67.89$}{\footnotesize$\pm0.02$}&$ 10.35${\footnotesize$\pm0.07$}&{\boldmath $213.95$}{\footnotesize$\pm0.08$}& $10.03${\footnotesize$\pm0.14$}&{\boldmath $80.19$}{\footnotesize$\pm0.06$}\\
    \bottomrule
    \end{tabular}
    \caption{KL and reconstruction part of ELBO averaged over test set by a single sample from posterior. }
    \label{tab:reconst_kl}
\end{table*}

\section{t-SNE visualization of Fashion MNIST latent space}
\vspace{-0.1cm}

We showed t-SNE visualization of MNIST latent space in the figure 5. Here we show the same plot for fashion-mnist. Interestingly, some classes are very close to each other (Pullover-shirt-dress) and transition between them happens very smoothly while some other classes are more separated.

\begin{table}[H]
    \centering
    \begin{tabular}{@{}c@{\hspace{0.015\linewidth}}c@{}}
        \includegraphics[width=4cm]{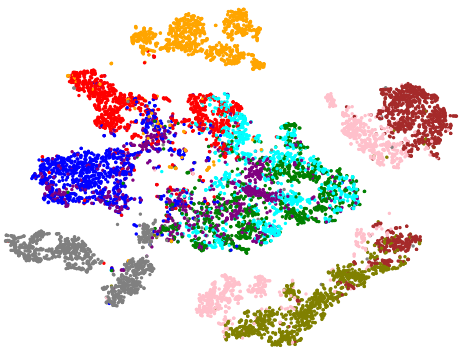} & \includegraphics[width=4cm]{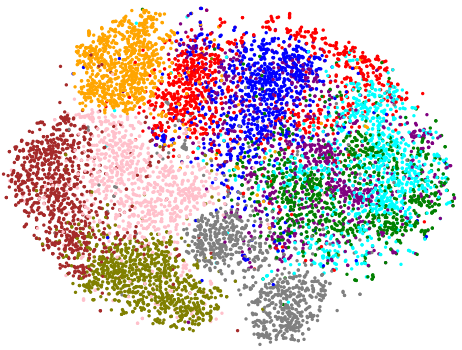}\\
        \scriptsize{Exemplar VAE on Fashion MNIST} & \scriptsize{VAE on Fashion MNIST} \\
    \end{tabular}
    \caption{\small t-SNE visualization of learned latent representations for Fashion-MNIST test points, colored by labels.
    \label{fig:tSNE-MNIST}}
\end{table}

\section{Experimental Details}

\subsection{Architectures}
All of the neural network architectures are based on the VampPrior of Tomczak \& Welling~\cite{tomczak2017vae}\footnote{https://github.com/jmtomczak/vae\_vampprior} except PixelSNAIL. We leave tuning the architecture of Exemplar VAEs to future work. To describe the network architectures, we follow the notation of LARS~\cite{bauer2018resampled}. Neural network layers used are either convolutional (denoted CNN) or fully-connected (denoted MLP), and the number of units are written inside a bracket separated by a dash (e.g., MLP[300-784] means a fully-connected layer with 300 input units and 784 output units). We use curly bracket to show concatenation. $d_z$ refers to the dimensionality of the latent space. \\\\
a) \textbf{VAE}:
\begin{eqnarray*}
    q_{\phi}(\z \mid \x) &=& \mathcal{N}(\z;~ \mu_{\z}(\x), \Lambda_{\z}(\x)) \\
    p_{\phi}(\x \mid \z) &=& \textrm{Bernoulli}(x, \mu_{\x}(\z) )\\
\end{eqnarray*}
\begin{eqnarray*}
    \textrm{Encoder}_{\z}(\x) &=& \textrm{MLP}~[784-300-300] \\
    \log \Lambda_{\z}^2(\x) &=& \textrm{MLP}[\textrm{Encoder}_{\z}(x)-d_{\z}]\\
    \mu_{\z}(\x) &=& \textrm{MLP}[\textrm{Encoder}_{\z}(x)-d_{\z}]\\
    \mu_{\x}(\z) &=& \textrm{MLP}[d_{\z}-300-300-784]
\end{eqnarray*}
\\
b) \textbf{HVAE}:
\begin{eqnarray*}
    q_{\phi}(\z_2 \mid \x) &=& \mathcal{N}(\z_2;~ \mu_{\z_2}(\x), \Lambda_{\z_2}(\x)) \\
    q_{\phi}(\z_1 \mid \x ,~ \z_2) &=& \mathcal{N}(\z_1;~ \mu_{\z_1}(\x,~ \z_2), \Lambda_{\z_1}(\x, ~ \z_2)) \\
    p_{\phi}(\z_1 \mid \z_2) &=& \mathcal{N}(\z_1;~ \hat{\mu}_{\z_1}(\z_2), \hat{\Lambda}_{\z_1}(\z_2))\\
    p_{\phi}(\x \mid \z_1, ~\z_2) &=& \textrm{Bernoulli}(\x, \mu_{\x}(\z_1,~ \z_2) )\\
\end{eqnarray*}
\begin{eqnarray*}
    \textrm{Encoder}_{\z_2}(\x) &=& \textrm{MLP}[784-300-300]\\
    \log \Lambda_{\z_2}^2(\x) &=& \textrm{MLP}[\textrm{Encoder}_{\z_2}(\x)-d_{\z_2}]\\
    \mu_{\z_2}(\x) &=& \textrm{MLP}[\textrm{Encoder}_{\z_2}(\x)-d_{\z_2}]\\
    \textrm{Encoder}_{\z_1}(\x, \z_2) &=& \textrm{MLP}[\{\textrm{MLP}[d_{\z_2}-300], \textrm{MLP}[784-300]\}-300]\\
    \log \Lambda_{\z_1}^2(\x, \z_2) &=& \textrm{MLP}[\textrm{Encoder}_{\z_1}(\x, \z_2)-d_{\z_1}]\\
    \mu_{\z_1}(\x, \z_2) &=& \textrm{MLP}[\textrm{Encoder}_{\z_1}(\x, \z_2)-d_{\z_1}]\\
    \textrm{Decoder}_{\z_1}(\z_2) &=& \textrm{MLP}[d_{\z_2}-300-300]\\
    \log \hat{\Lambda}_{\z_1}^2(\z_2) &=& \textrm{MLP}[\textrm{Decoder}_{\z_1}(\z_2)-d_{\z_1}]\\
    \hat{\mu}_{\z_1}(\z_2) &=& \textrm{MLP}[\textrm{Decoder}_{\z_1}(\z_2)-d_{\z_1}]\\
    \mu_{\x}(\z_1, \z_2) &=& MLP[\{\textrm{MLP}[d_{\z_1}-300], \textrm{MLP}[d_{\z_2}-300]\}-300-784]\\
\end{eqnarray*}
c) \textbf{ConvHVAE}:
The generative and variational posterior distributions are identical to HVAE.
\begin{eqnarray*}
    \textrm{Encoder}_{\z_2}(\x) &=& \textrm{CNN}[28\times28\times1-32\times32\times32-12\times12\times32-12\times12\times64-7\times7\times64\\&&-7\times7\times6]\\
    \log \Lambda_{\z_2}^2(\x) &=& \textrm{MLP}[\textrm{Encoder}_{\z_2}(\x)-d_{\z_2}]\\
    \mu_{\z_2}(\x) &=& \textrm{MLP}[\textrm{Encoder}_{\z_2}(\x)-d_{\z_2}]\\
    \textrm{ConvEncoder}_{\z_1}(\x) &=& \textrm{CNN}[28\times28\times1-32\times32\times32-12\times12\times32-12\times12\times64-7\times7\times64-7\times7\times6]\\
    \textrm{Encoder}_{\z_1}(\x, \z_2) &=& \textrm{MLP}[\{\textrm{MLP}[d_{\z_2}\\&&-7\times7\times6], \textrm{ConvEncoder}_{\z_1}(\x)\}-300]\\
    \log \Lambda_{\z_1}^2(\x, \z_2) &=& \textrm{MLP}[\textrm{Encoder}_{\z_1}(\x, \z_2)-d_{\z_1}]\\
    \mu_{\z_1}(\x, \z_2) &=& \textrm{MLP}[\textrm{Encoder}_{\z_1}(\x, \z_2)-d_{\z_1}]\\
    \textrm{Decoder}_{\z_1}(\z_2) &=& \textrm{MLP}[d_{\z_2}-300-300]\\
    \log \hat{\Lambda}_{\z_1}^2(\z_2) &=& \textrm{MLP}[\textrm{Decoder}_{\z_1}(\z_2)-d_{\z_1}]\\
    \hat{\mu}_{\z_1}(\z_2) &=& \textrm{MLP}[\textrm{Decoder}_{\z_1}(\z_2)-d_{\z_1}]\\
    \textrm{MLPDecoder}_{\x}(\z_1, \z_2) &=& \textrm{MLP}[\{\textrm{MLP}[d_{\z_1}-300], \textrm{MLP}[d_{\z_2}-300]\}-784]\\
    \textrm{ConvDecoder}_{\x}&=&\textrm{CNN}[28\times28\times64-28\times28\times64-28\times28\times64-28\times28\times64-28\times28\times1]\\
    \mu_{\x}(\z_1, \z_2) &=& [\textrm{MLPDecoder}_{\x}(\z_1, \z_2)-\textrm{ConvDecoder}_{\x}]\\
\end{eqnarray*}

d) \textbf{PixelSNAIL HVAE}:
The generative and variational posterior distributions are identical to HVAE.
\begin{eqnarray*}
    \textrm{Encoder}_{\z_2}(\x) &=& \textrm{CNN}[28\times28\times1-32\times32\times32-12\times12\times32-12\times12\times64\\ &&- 7\times7\times64-7\times7\times6]\\
    \log \Lambda_{\z_2}^2(\x) &=& \textrm{MLP}[\textrm{Encoder}_{\z_2}(\x)-d_{\z_2}]\\
    \mu_{\z_2}(\x) &=& \textrm{MLP}[\textrm{Encoder}_{\z_2}(\x)-d_{\z_2}]\\
    \textrm{ConvEncoder}_{\z_1}(\x) &=& \textrm{CNN}[28\times28\times1-32\times32\times32-12\times12\times32-12\times12\times64\\
    &&-7\times7\times64-7\times7\times6]\\
    \textrm{Encoder}_{\z_1}(\x, \z_2) &=& \textrm{MLP}[\{\textrm{MLP}[d_{\z_2}-7\times7\times6], \textrm{ConvEncoder}_{\z_1}(\x)\}-300]\\
    \log \Lambda_{\z_1}^2(\x, \z_2) &=& \textrm{MLP}[\textrm{Encoder}_{\z_1}(\x, \z_2)-d_{\z_1}]\\
    \mu_{\z_1}(\x, \z_2) &=& \textrm{MLP}[\textrm{Encoder}_{\z_1}(\x, \z_2)-d_{\z_1}]\\
    \textrm{Decoder}_{\z_1}(\z_2) &=& \textrm{MLP}[d_{\z_2}-300-300]\\
    \log \hat{\Lambda}_{\z_1}^2(\z_2) &=& \textrm{MLP}[\textrm{Decoder}_{\z_1}(\z_2)-d_{\z_1}]\\
    \hat{\mu}_{\z_1}(\z_2) &=& \textrm{MLP}[\textrm{Decoder}_{\z_1}(\z_2)-d_{\z_1}]\\
    \textrm{MLPDecoder}_{\x}(\z_1, \z_2, \x) &=& \{\textrm{MLP}[d_{\z_1}-784], \textrm{MLP}[d_{\z_2}-784],\x\}\\
    \textrm{AutoRegressiveDecoder}_{\x}&=&[\textrm{ResNet-MaskedCNN}[28\times28\times64] \times 4\\&& -\textrm{Self-Attention}-\textrm{MaskedCNN}[28\times28\times1]]\\
    \mu_{\x}(\z_1, \z_2) &=& [\textrm{MLPDecoder}_{\x}(\z_1, \z_2, \x)-\textrm{AutoRegressiveDecoder}_{\x}]\\
\end{eqnarray*}

e) \textbf{CelebA Architecture}:
\begin{eqnarray*}
    q_{\phi}(\z \mid \x) &=& \mathcal{N}(\z;~ \mu_{\z}(\x), \Lambda_{\z}(\x)) \\
    p_{\phi}(\x \mid \z) &=& \textrm{Discretized\_Logistic}(x, \mu_{\x}(\z), \sigma^2)\\
\end{eqnarray*}
\begin{eqnarray*}
    \textrm{Encoder}_{\z}(\x) &=& \textrm{CNN}~[64\times64 \times 3 - 32\times32\times64 - 16\times16\times128 - 8\times8\times256 - 4\times4\times512] \\
    \log \Lambda_{\z}^2(\x) &=& \textrm{MLP}[\textrm{Encoder}_{\z}(x)-d_{\z}]\\
    \mu_{\z}(\x) &=& \textrm{MLP}[\textrm{Encoder}_{\z}(x)-d_{\z}]
     \\ \mu_{\x}(\z) &=& \textrm{CNN}[8\times8\times512-16\times16\times256 - 32\times32\times128 - 64\times64\times64 - 64\times64\times3]
\end{eqnarray*}

As the activation function, the gating mechanism of \cite{dauphin2017language} is used throughout. So for each layer we have two parallel branches where the sigmoid of one branch is multiplied by the output of the other branch. In ConvHVAE the kernel size of the first layer of $\textrm{Encoder}_{\z_2}(x)$ is 7 and the third layer used kernel size of 5. The last layer of $\textrm{ConvDecoder}_{\x}$ used kernel size of 1 and all the other layers used $3\times3$ kernels. For CelebA we used kernel size of 5 for each layer and combination of batch norm and ELU activation after each convolution layer.
\vspace*{-0.1cm}
\subsection{Hyper-parameters}
\vspace*{-0.1cm}

We use Graident Normalized Adam \cite{yu2017normalized} with Learning rate of $5e-4$ and minibatch size of $100$ for all of the datasets. For gray-scale datasets We dynamically binarize each training data, but we do not binarize the exemplars that serve as the prior. 
We utilize early stopping for training VAEs, where we stopped the training if for $50$ consecutive epochs the validation ELBO does not improve. We use 40 dimensional latent spaces for gray-scale datasets while using 128 dimensional latent for CelebA. To limit the computation costs of convolutional architectures, we considered kNN based on euclidean distance in the latent space, where $k$ set to $10$ for gray-scale datasets and $5$ for CelebA. The number of exemplars set to the half of the training data except in the ablation study section. 

\section{Misclassified MNIST Digits }
A classifier trained using exemplar augmentation reached average error of $0.69\%$. Here we show the test examples misclassified. 
\begin{figure}[H]
\centering
\includegraphics[width=0.35\linewidth]{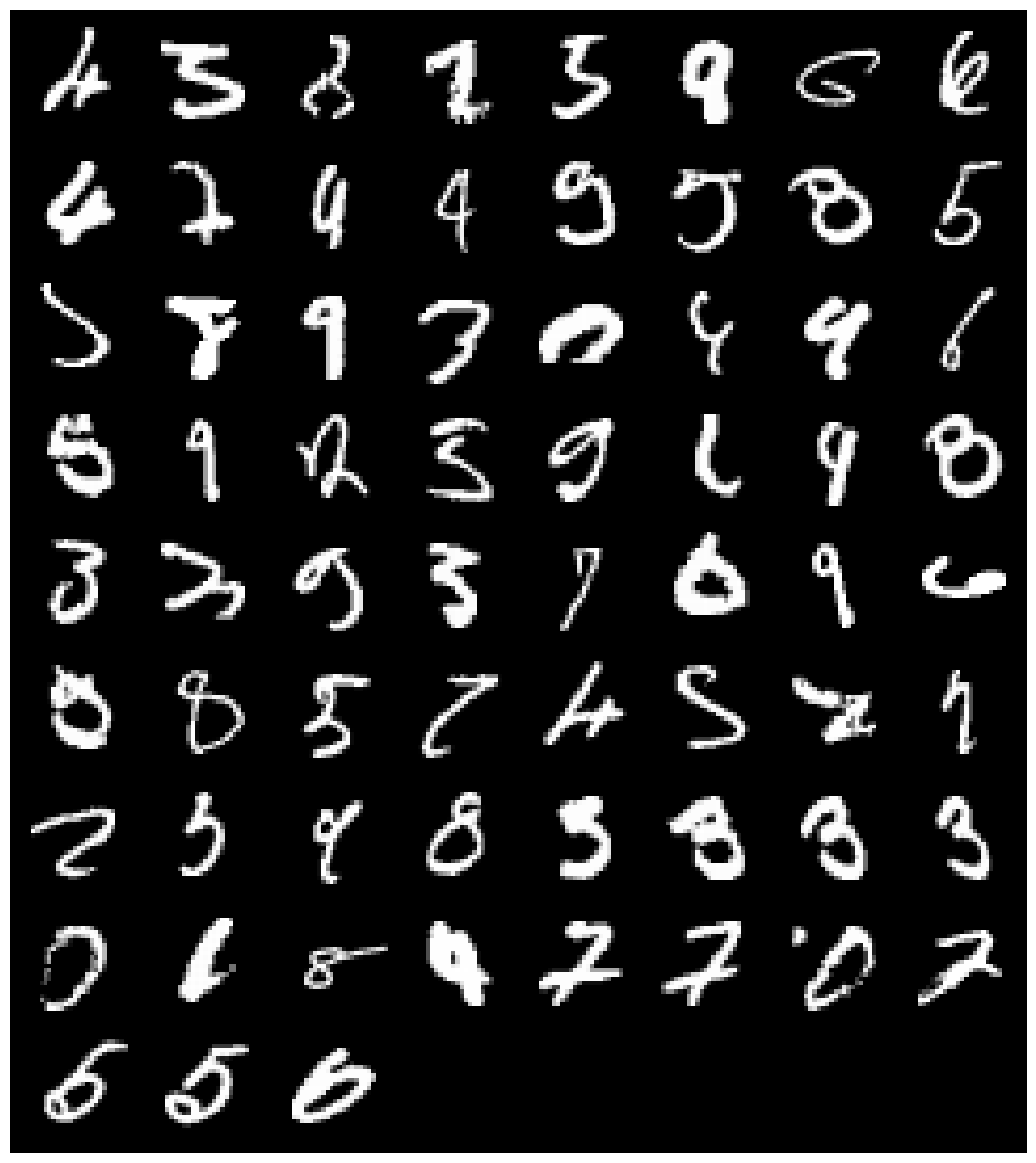}
\caption{Misclassified images from MNIST test set for a two layer MLP trained with Exemplar VAE augmentation.}
\label{fig:augmentation}
\end{figure}

\end{document}